\PassOptionsToPackage{dvipsnames, table, xcdraw}{xcolor}
\documentclass{vivo}
\usepackage{booktabs}
\usepackage{amsmath, amssymb, mathtools, bm}
\usepackage{multirow, enumitem, float, wrapfig}

\usepackage{graphicx} 
\usepackage{newfloat}
\usepackage{listings}
\usepackage{booktabs} 
\usepackage{multirow} 
\usepackage{caption}
\usepackage{subcaption}
\usepackage[numbers]{natbib}
\usepackage{cleveref}
\usepackage{pdfpages}

\usepackage{hyperref}
\usepackage{url}
\usepackage{booktabs}
\usepackage{makecell}
\usepackage{graphicx}
\usepackage{capt-of}
\usepackage{multirow}

\PassOptionsToPackage{hyphens}{url} 
\usepackage{url}

\usepackage[utf8]{inputenc}
\usepackage{tabularx}

\newcommand{\nMethod}{Hear the World in Stereo}

\title{\nMethod{}: Learning Dynamic Spatial Correspondence for Immersive Joint Video-Audio Generation}

\author[]{Hanmo Chen$^{1,2*}$} 
\author[]{Chengcheng Liu$^{2*}$}
\author[]{Tianxiao Chen$^{2}$}
\author[]{Zheyu Zhang$^{2}$} 
\author[]{Siming Zheng$^{2}$} 
\author[]{Jinwei Chen$^{2}$} 
\author[]{Xu Yang$^{1\dagger}$} 
\author[]{Cheng Deng$^{1}$} 
\author[]{Bo Li$^{2}$}
\author[]{Peng-Tao Jiang$^{2\dagger}$}

\affiliation[]{$^{1}$Xidian University} 

\affiliation[]{ \\ $^{2}$vivo BlueImage Lab, vivo Mobile Communication Co., Ltd.}

\affiliation[]{ \\ $\dagger$: Corresponding authors.}
\affiliation[]{\\ $^*$ Equal contribution.}

\abstract{
Recent joint video-audio generation models have achieved strong semantic correspondence and temporal synchronization. However, applications such as AR/VR and interactive gaming further require stereo audio to provide an immersive sense, which remains largely overlooked. Effective stereo audio requires the perceived sound location to evolve consistently with the motion of its corresponding visual source. We refer to this property as \emph{Dynamic Spatial Correspondence} and propose StereoBind, a framework that binds visual source motion to stereo sound generation. StereoBind uses motion tracks to coordinate visual motion and stereo audio through three complementary mechanisms. Visual Motion Binding establishes source-aware audiovisual correspondence, the Spatial Track Encoder captures absolute source positions, and Residual Track RoPE models relative motion. For supervision and evaluation, we construct StereoWorld-29K, a large-scale stereo audio-video dataset with paired motion tracks, and StereoWorldBench for measuring audiovisual spatial consistency. Experiments show that StereoBind substantially improves spatial alignment in stereo audio generation over existing models while preserving overall audiovisual quality.
}

\projectpage{https://vivocameraresearch.github.io/Stereo-Bind-Project/}

\begin{document}

\maketitle

\begin{figure}[t]
    \centering
    \includegraphics[page=1, width=\linewidth]{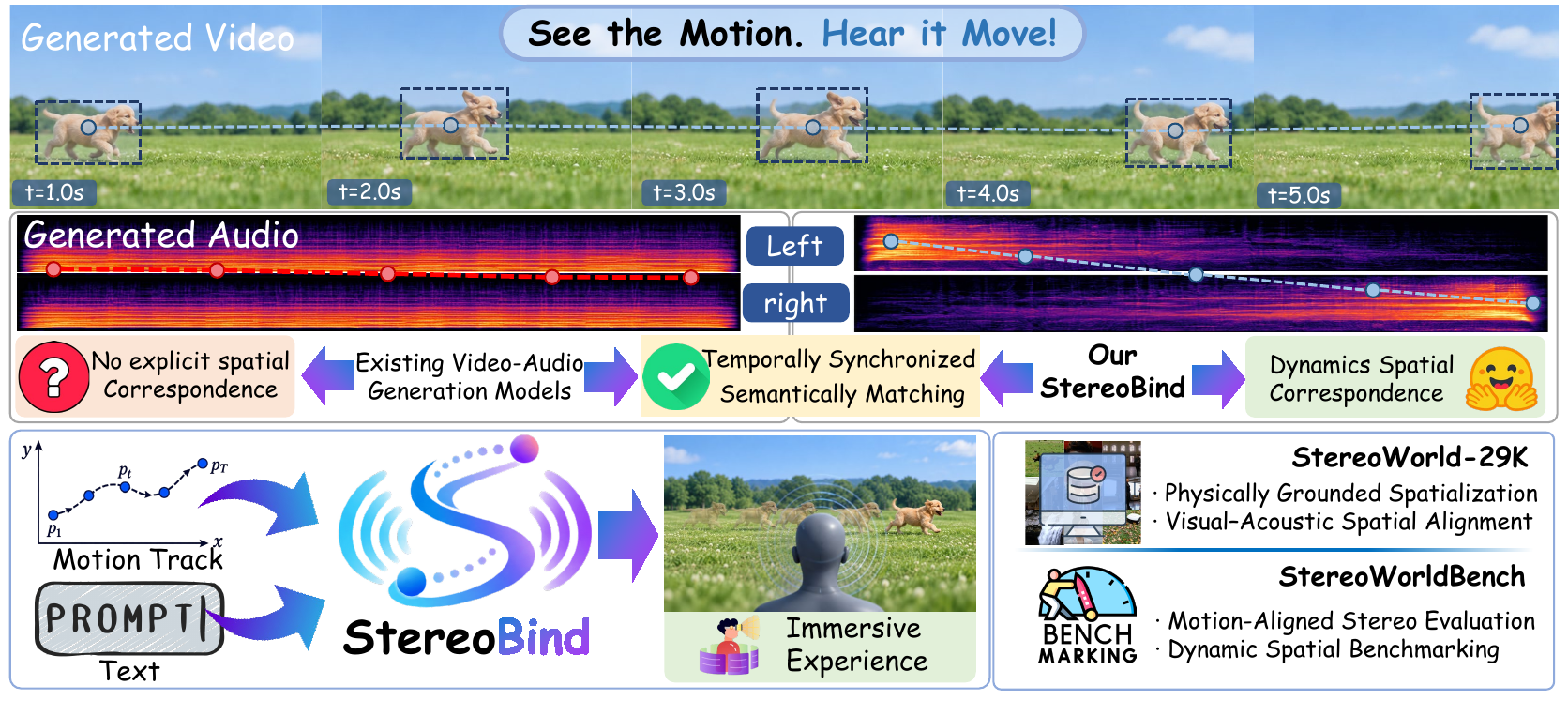}
    \caption{To the best of our knowledge, StereoBind is the first framework for track-conditioned stereo VA generation, with StereoWorld-29K as the first large-scale dataset and StereoWorldBench as the first benchmark for this task.}
\end{figure}

\section{Introduction} \label{sec:intro}

Joint video-audio (VA) generation is rapidly evolving toward coherent audiovisual synthesis. Recent foundation models such as Seedance~\citep{seedance2026seedance}, LTX-2~\citep{hacohen2026ltx}, MOVA~\citep{team2026mova}, and Minimax-H3 focus primarily on \emph{what} is heard through semantic correspondence and \emph{when} it occurs through temporal synchronization. However, VR/AR and interactive games demand not only semantic and temporal coherence but also emphasis immersive entertainment~\citep{zhu2025asaudio}. Users should hear sounds from directions consistent with visible sources, with perceived sound locations evolving as the sources move~\citep{wang2024hearing}. This motivates us to investigate \emph{where} sounds originate and how their locations evolve over time, a property we term \emph{dynamic spatial correspondence}, which remains largely underexplored in existing methods.

Although existing VA models~\citep{hacohen2026ltx, low2025ovi} support two-channel audio generation, such outputs do not guarantee meaningful stereophonic spatial cues. In stereo audio, spatial perception arises from structured differences between the left and right channels~\citep{liu2026prismaudio}. This principle mirrors human auditory perception, in which differences between the signals received by the two ears provide cues about the location of a sound source. In particular, horizontal sound localization relies strongly on interaural time differences (ITDs) and interaural level differences (ILDs)~\citep{blauert1996spatial}, which encode disparities in arrival time and intensity between the two ears, enabling listeners to infer the direction of a sound source. As a sound source moves, these binaural cues should evolve consistently with the corresponding motion. We refer to this consistency between sound source motion and perceived auditory location over time as dynamic spatial correspondence. Establishing dynamic spatial correspondence is challenging because a visual motion track cannot be directly translated into stereo audio. The motion track explicitly describes where a source moves in the video, while auditory location is reflected indirectly through spatial cues such as ITDs and ILDs. Bridging these heterogeneous representations requires establishing a link between the visual source and its associated auditory cues~\citep{dagli2025see,zhang2025isdrama}.

To address these challenges, we introduce StereoBind, which uses the motion track to condition both video and stereo audio generation. StereoBind comprises three complementary components. Visual Motion Binding (VMB) serve as a cross-modal representation bridge that aggregates the motion track with the corresponding visual entity and transforms this information into a spatially informative representation. The resulting representation is injected into the audio stream, allowing visual motion to directly guide the spatial structure of the generated audio. A Spatial Track Encoder (STE) maps the motion track into spatial track embeddings, which are then used to modulate the audio latents, providing explicit absolute spatial conditioning for audio generation. Residual Track RoPE (RT-RoPE) incorporates relative track displacement, providing relative spatial conditioning. Together, these components enable spatially coherent audio generation aligned with visual motion.

Training StereoBind to capture dynamic spatial correspondence requires VA data with reliable spatial supervision. Existing large-scale datasets such as VGGSound~\citep{chen2020vggsound} and OpenHumanVid~\citep{li2025openhumanvid} primarily support semantic and temporal correspondence, but rarely provide reliable stereo cues. Moreover, collecting large-scale real-world data with calibrated spatial recording equipment is also costly. We therefore construct \textbf{StereoWorld-29K} through two complementary pipelines. A Visual-Audio Spatialization (VAS) Pipeline spatializes audio according to the trajectories of sound sources in real videos, while a Spatial Data Synthesis (SDS) Pipeline generates track-controlled scenes with diverse configurations. Together, these two pipelines provide scalable and interpretable supervision between sound source motion and stereo audio. 

Beyond training supervision, spatial alignment also requires dedicated evaluation. We therefore introduce StereoWorldBench (SWBench), which measures the alignment between generated stereo audio and source motion to complement conventional audiovisual quality metrics. Experiments validate the effectiveness of our method. Our key contributions are summarized as follows:

\begin{itemize}
	\item To the best of our knowledge, we present the first framework for track-conditioned stereo VA generation, modeling the consistency between source motion and acoustic location.
	\item  We construct StereoWorld-29K, the first large-scale stereo VA dataset with explicit spatial supervision, and introduce StereoWorldBench to evaluate dynamic spatial correspondence.
    \item  We propose StereoBind, which models source binding, absolute spatial position, and relative motion through VMB, STE, and RT-RoPE, respectively.
\end{itemize}

\section{Related Work}

\subsection{Audio-Video Joint Generation}

VA generation has evolved from diffusion models~\citep{rombach2022high} toward Diffusion Transformer (DiT)~\citep{peebles2023scalable}. Early approaches such as MM-Diffusion~\citep{ruan2023mm} couple audio and video through cross-modal attention, while AV-DiT~\citep{wang2024av} adapts a pretrained image DiT with lightweight VA modules. More recently, Ovi~\citep{low2025ovi}, LTX-2~\citep{hacohen2026ltx}, and MOVA~\citep{team2026mova} adopt dual-stream architectures to jointly model audio and video, while UniAVGen~\citep{zhang2026uniavgen} further incorporates face-aware modulation. Beyond architectural design, JavisDiT~\citep{liu2026javisdit} and Harmony~\citep{hu2026harmony} strengthen audiovisual correspondence through mechanisms such as spatiotemporal priors, modality-specific modeling, and synchronization-aware training~\citep{liu2026javisdit, hu2026harmony}. Other conditional generation methods, including  Syncphony~\citep{song2026syncphony}, DreamID-Omni~\citep{guo2026dreamid} and Hallo-Live~\citep{li2026hallo}, further explore explicit cross-modal conditioning and binding. These advances improve semantic correspondence and temporal synchronization, yet the alignment between motion and auditory location remains largely unexplored.

\subsection{Stereo Audio Generation}

Stereo audio research has increasingly focused on controllable and scene-aware synthesis~\citep{zhu2025asaudio}. A growing body of work incorporates spatial attributes as explicit generation conditions. SpatialSonic~\citep{sun2025both} models directional states for stereo synthesis, while ISDrama~\citep{zhang2025isdrama} extends such control to multi-speaker speech. Visual information has also been exploited to guide stereo audio generation. ViSAGe~\citep{kim2025visage} generates directional First-Order Ambisonics (FOA) audio from visual content, and FoleyDesigner~\citep{li2026foleydesigner} couples scene analysis with controllable stereo Foley generation. Recent methods incorporate explicit scene geometry, with SonoWorld~\citep{jin2026sonoworld} constructing spatially consistent 3D audiovisual scenes and Sonic4D~\citep{xie2026sonic4d} combining source localization and physics-based spatialization. In parallel, OWL~\citep{biswas2026owl} and BAT~\citep{zheng2024bat} explore stereo audio understanding with large language models. Despite these advances, dynamic spatial correspondence between sound source motion and auditory spatial cues remains underexplored in VA generation.

\begin{figure*}[t]
  \centering
  \includegraphics[width=\linewidth]{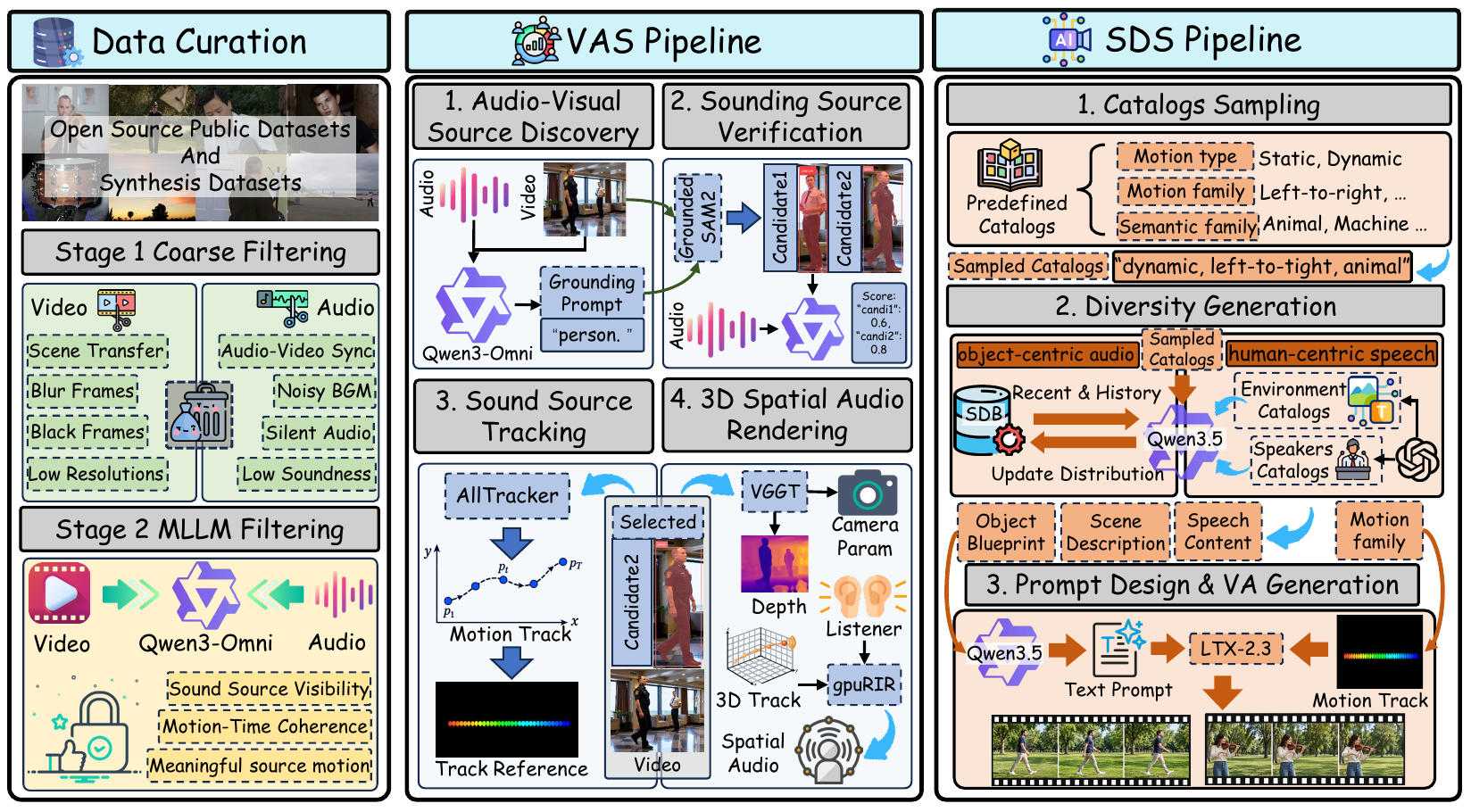}
  \caption{\textbf{Construction pipeline of StereoWorld-29K.} We combine curated real-world videos with synthetic audiovisual data. The VAS Pipeline localizes and tracks sound sources and renders track-guided stereo audio, while the SDS Pipeline generates diverse track-controlled scenes.}
\label{fig:datasets}
\end{figure*}

\section{StereoWorld-29K Dataset}

StereoWorld-29K consists of two data branches, with the overall construction pipeline shown in Figure~\ref{fig:datasets}. The real-world branch draws from large-scale public datasets, preserving natural audiovisual content but offering limited motion diversity. The synthetic branch generates diverse audiovisual scenes with controllable source motion to support stereo audio generation. Reliable sound-source trajectories are required to provide spatial supervision for both branches. However, existing Audio-Visual Segmentation (AVS) models trained on curated datasets often generalize poorly to unconstrained scenes. We therefore develop the VAS Pipeline to robustly localize and track sound sources for stereo rendering. Detailed statistics and analysis are provided in Appendix~\ref{sec:appendix_stereoworld}.

\subsection{Data Curation}

We collect videos from VGGSound~\citep{chen2020vggsound}, OpenHumanVid~\citep{li2025openhumanvid}, FoleyBench~\citep{dixit2026foleybench}, and LU-AVS~\citep{liu2024benchmarking}, and apply a two-stage filtering pipeline. First, we filter out low-quality samples based on video duration and resolution, audio decodability, silence ratio, and signal-to-noise ratio. Second, we use Qwen3-Omni~\citep{xu2025qwen3} for motion-aware semantic filtering. We retain clips in which the audio originates from a spatially identifiable source whose motion is temporally coherent relative to the camera. This process focuses the dataset on dynamic sound-source motion relevant to dynamic spatial correspondence.

\subsection{Visual-Audio Spatialization Pipeline}

Given an audiovisual clip, we first use Qwen3-Omni~\citep{xu2025qwen3} to identify potential visible sound sources and generate grounding prompts, which are then passed to Grounded SAM~\citep{ren2024grounded} to obtain masks and bounding boxes. Since grounding may return multiple candidate instances, we perform candidate-level audiovisual verification. For each candidate, we construct a candidate-centric visual clip and pair it with the original audio, allowing Qwen3-Omni~\citep{xu2025qwen3} to assess whether the candidate is consistent with the audible content. The candidate with the strongest audiovisual correspondence is selected, after which we use AllTracker~\citep{harley2025alltracker} to track the selected source throughout the video and obtain its motion track. 

We then use VGGT~\citep{wang2025vggt} to estimate scene depth and camera parameters. By combining the 2D source track with its estimated depth and camera intrinsics, we back-project the source into a camera-centered 3D coordinate system. Treating the camera as a virtual listener, we place left and right virtual receivers with a fixed interaural separation and use gpuRIR~\citep{diaz2021gpurir} to compute a sequence of Room Impulse Responses (RIRs) along the source track. Convolving these RIRs with the mono source audio produces stereo signals whose spatial cues evolve consistently with the modeled source geometry and room acoustics. In this way, the VAS Pipeline provides explicit, geometry-aware spatial supervision for stereo video-audio generation.

\subsection{Spatial Data Synthesis Pipeline}

Our SDS Pipeline follows a two-stage design. Stage~1 emphasizes diversity by stochastically sampling diverse spatial scene specifications, while Stage~2 converts these specifications into generation-ready prompts for VA synthesis.

In Stage~1, we first sample a stereo configuration that defines the motion and semantic attributes of each synthetic example, including motion type (static or dynamic), motion family, and semantic family. Qwen3.5~\citep{yang2025qwen3} generates and translated a corresponding motion track into a textual description. Subsequently, we instantiate this configuration through two complementary branches: an object-centric branch and a human-centric branch. In the object-centric branch, Qwen3.5~\citep{yang2025qwen3} takes the sampled configuration together with statistics from a continuously updated Scene Distribution Bank (SDB) to generate a structured blueprint describing the source appearance, sound, motion, and scene context. The SDB maintains statistics over previously accepted scene specifications to reduce repetition in subsequent samples. In the human-centric branch, we first use GPT-5.6 to construct a predefined catalog comprising 47 speaker profiles and 41 concrete environments. Given the identity of the sampled speaker profile, action, and scene context, Qwen3.5 then generates a natural spoken sentence tailored to the scene. 

In stage~2, the branch-specific outputs are then passed to separate Qwen3.5 prompt generators, which convert them into prompts for VA models. We use a higher sampling temperature in Stage~1 to encourage semantic and scene diversity, and a lower temperature in Stage~2 to faithfully preserve the details of the planned scene specifications during prompt generation.

\begin{figure*}[t]
  \centering
  \includegraphics[width=\linewidth]{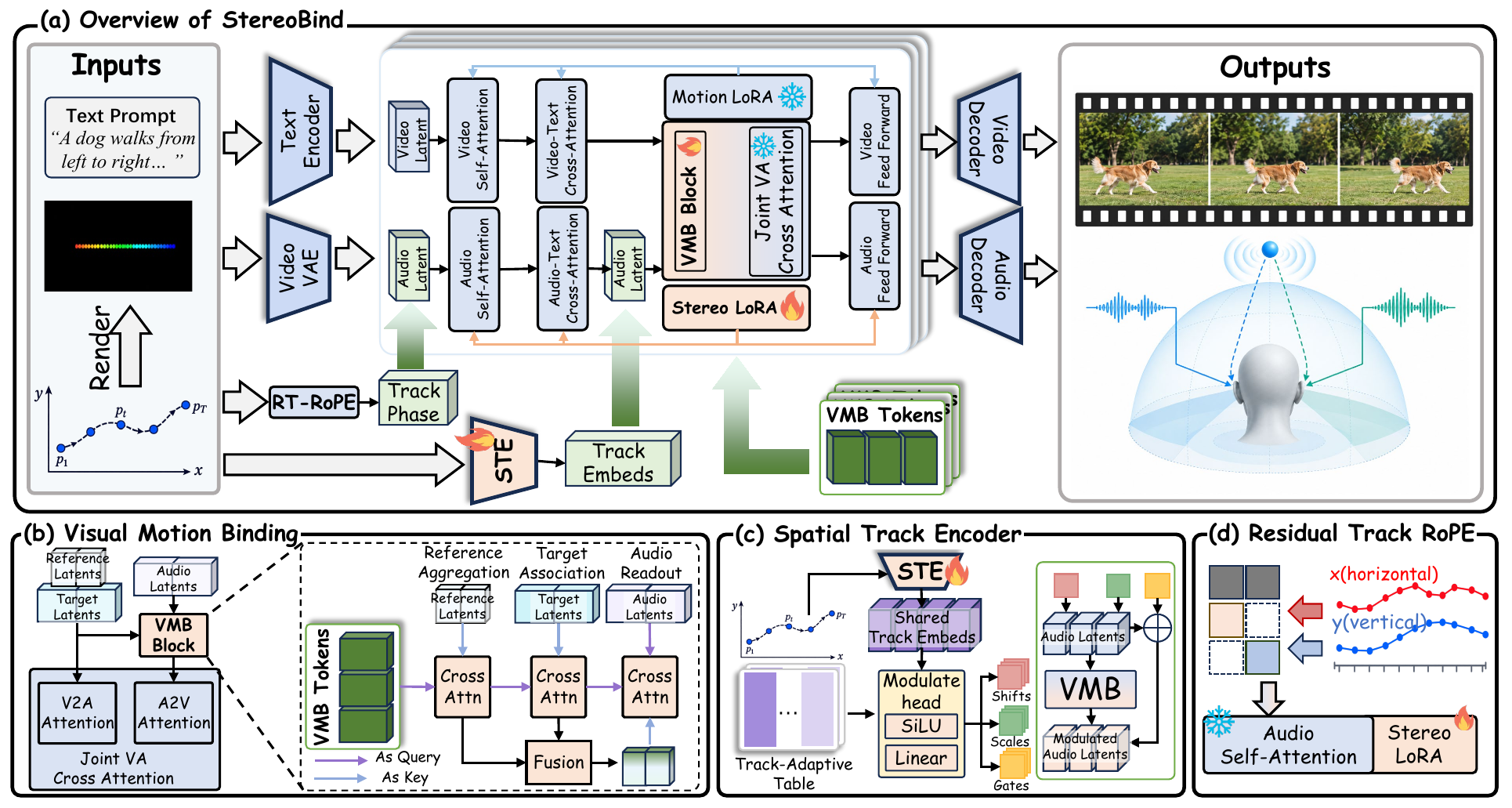}
  \caption{\textbf{Overview of the StereoBind framework.} StereoBind introduces VMB for audiovisual source binding, STE for absolute track conditioning, and RT-RoPE for relative motion modeling.}
\label{fig:methods}
\end{figure*}

\section{Method}

\subsection{Problem Formulation and Overview}

Given a text description $c$ and a normalized motion track with $T_p$ frames $P=(\mathbf p_i)_{i=1}^{T_p}$, where $\mathbf p_i=(x_i,y_i)\in[0,1]^2,$, we model the conditional joint distribution

\begin{equation}
p_\theta(V,\mathbf A\mid c,P,\mathcal{R}(P)),
\end{equation}

where $V$ denotes the generated video, $\mathbf{A}=(A^L,A^R)$ the corresponding stereo audio, and $R_P=\mathcal{R}(P)$ the visual reference rendered from the motion track $P$. Our goal is to enforce dynamic spatial correspondence, such that the visible sound-producing source in $V$ follows the prescribed track $P$, while the perceived sound location in $\mathbf{A}$ evolves consistently with the resulting visual source motion. As illustrated in Figure~\ref{fig:methods}(a), StereoBind builds on the LTX-2.3~\citep{hacohen2026ltx} backbone and consists of three complementary components. VMB bridges the visual and audio modalities by transferring source-aware motion context to the audio stream. STE provides absolute spatial conditioning, while RT-RoPE models relative motion with a trainable Stereo LoRA for spatial adaptation. Together, these components enable stereo audio generation aligned with visual source motion without explicitly estimating ITDs or ILDs.

\subsection{Visual Motion Binding}

The motion reference specifies \emph{how} the source should move, whereas the evolving target-video state provides the visual context of \emph{what} is being generated. Neither alone explicitly establishes dynamic spatial correspondence between visual source motion and perceived sound location. We therefore introduce Visual Motion Binding (VMB), consisting of $N_\text{VMB}=64$ learnable VMB Tokens in each VMB Blocks. As illustrated in Figure~\ref{fig:methods}(b), the VMB Block is placed before the Joint VA Attention in each transformer block and progressively integrates the motion reference, target-video state, and audio stream through a \emph{reference--target--audio} pathway. Starting from learnable tokens $S_0^\ell$ at blcok $\ell$, the VMB Block first aggregates the motion-reference representation $X_{\mathrm{ref}}^\ell$ to encode the prescribed track. The resulting tokens then attend to the current target-video representation $X_{\mathrm{tar}}^\ell$, contextualizing the prescribed motion with the visual entity and scene being synthesized. We further retain the residual $S_t^\ell-S_r^\ell$ to characterize how the target-video context modifies the reference-conditioned representation. The audio stream then attends to the fused representation through cross-attention:

\begin{equation}
\begin{aligned}
S_r^\ell
&= S_0^\ell
+ \mathcal{A}_\ell^r\!\left(
\operatorname{LN}(S_0^\ell),
\operatorname{LN}(X_{\mathrm{ref}}^\ell)
\right), \\
S_t^\ell
&= S_r^\ell
+ \mathcal{A}_\ell^t\!\left(
\operatorname{LN}(S_r^\ell),
\operatorname{LN}(X_{\mathrm{tar}}^\ell)
\right), \\
S^\ell
&= \mathcal{F}_\ell\!\left(
[S_r^\ell, S_t^\ell, S_t^\ell - S_r^\ell]
\right), \\
D_s^\ell
&= \mathcal{A}_\ell^a\!\left(
\operatorname{LN}({X}_a^\ell),
\operatorname{LN}(S^\ell)
\right).
\end{aligned}
\end{equation}

Here, $\mathcal A_\ell^r$, $\mathcal A_\ell^t$, and $\mathcal A_\ell^a$ denote cross-attention for reference aggregation, target association, and audio readout, respectively, while $\mathcal F_\ell$ denotes a MLP projection applied to the concatenated features $[\cdot]$. $X_{\mathrm{ref}}^\ell$ represents the motion-reference features, $X_{\mathrm{tar}}^\ell$ denotes the video state during denoising, and $X_a^\ell$ denotes the audio representation at block $\ell$. Finally, $D_s^\ell$ denotes the spatially conditioned audio feature produced by the VMB Block, which is subsequently fed into the Joint VA Cross-Attention.

\subsection{Spatial Track Encoder}

VMB provides source-aware visual context but lacks token-wise temporal spatial alignment. Consequently, we introduce a STE shared across Transformer blocks that maps the motion track $P$ into a temporally structured representation. Since audio representations evolve across Transformer layers, directly applying the same track embedding at every block may introduce a representation mismatch. We therefore further introduce layer-specific Track-Adaptive modulation to adapt the shared track representation to each block. As shown in Figure~\ref{fig:methods}(c), given $P\in\mathbb{R}^{T_p\times2}$, the overall process is

\begin{equation}
\begin{gathered}
H
=
\operatorname{LN}\!\left(
\operatorname{TemporalTransformer}\!\left(
\operatorname{Linear}_p(P)
\right)
\right),\\
[\beta_\ell,\gamma_\ell,g_\ell]
=
\mathcal{U}_3\!\left(
\operatorname{Linear}_u(\operatorname{SiLU}(H))
+
B_\ell
\right).
\end{gathered}
\end{equation}

The shared representation $H$ captures temporal motion context. At block $\ell$, the learnable Track-Adaptive table $B_\ell\in\mathbb{R}^{1\times 3D_a}$ adapts $H$ to the layer-specific audio representation. The operator $\mathcal{U}3(\cdot)$ splits the feature dimension into three equal parts, yielding shift $\beta\ell$, scale $\gamma_\ell$, and gate $g_\ell$, where $\beta_\ell$ and $\gamma_\ell$ modulate the audio queries for VMB retrieval, while $g_\ell$ controls the strength of the retrieved update. Here, ${X}_a^\ell$ denotes the audio representation at block $\ell$ before VMB.

\begin{equation}
\begin{gathered}
\widetilde{X}_a^\ell
=
\operatorname{RMSNorm}(X_a^\ell)
\odot (1+\gamma_\ell)
+\beta_\ell, \\
X_{a,+}^\ell
=
\widetilde{X}_a^\ell
+
g_\ell \odot D_s^\ell .
\end{gathered}
\end{equation}

This design confines track conditioning to the VMB pathway while retaining the pretrained audio residual pathway and injecting track-dependent information through a gated residual update. The motion track thus governs how source-aware context is queried and incorporated into the audio representation. Together, STE and Track-Adaptive modulation bridge explicit motion tracks and latent audio representations.

\subsection{Residual Track RoPE}

The STE provides absolute spatial conditioning, while dynamic spatial correspondence further requires relative source motion. We therefore introduce Residual Track RoPE (RT-RoPE), which augments the pretrained temporal RoPE with track-dependent residual phases. As illustrated in Figure~\ref{fig:methods}(d), given a motion track $P$, we linearly resample it to the audio-token length and normalize it as $\bar P=\bigl((\bar x_j,\bar y_j)\bigr)_{j=1}^{N_a}$, where $(\bar x_j,\bar y_j)\in[-1,1]^2$. For each attention head, the rotary pairs are divided evenly into horizontal and vertical subsets $\mathcal I_x$ and $\mathcal I_y$. Let $n_{rp}$ denote the number of rotary pairs per head. For each spatial axis $d\in{x,y}$, we construct a logarithmically spaced rotary-frequency grid as $\omega_k^d=\frac{\pi}{2}\theta_{\mathrm{track}}^{\frac{k}{n_{rp}/2-1}}$, where $k=0,\ldots,n_{rp}/2-1$. We set $\theta_{\mathrm{track}}=2$ to provide a compact low-frequency spatial spectrum that matches the normalized track range and avoids overly rapid phase variation. The residual phase is defined as

\begin{equation}
\phi_{hjr}=
\begin{cases}
\tanh(a_x)\,\omega_r^x\,\bar x_j,
& r\in\mathcal I_x,\\
\tanh(a_y)\,\omega_r^y\,\bar y_j,
& r\in\mathcal I_y.
\end{cases}
\end{equation}

Here, $a_x$ and $a_y$ are learnable parameters controlling the strength of modulation, while $r$ denotes the local rotary-pair index within each subset. The $\tanh$ parameterization bounds the modulation coefficients, preventing excessive perturbation of the pretrained RoPE. The residual phase is added to the original RoPE as $\theta_{hjr}=\theta_{hjr}^{\mathrm{time}}+\phi_{hjr}$. Since RT-RoPE modifies the rotary geometry applied to the attention queries and keys, we augment the audio self-attention with a lightweight LoRA adapter while keeping the pretrained weights frozen. The adapter is jointly optimized with RT-RoPE, allowing the attention projections to adapt to the modified positional geometry. This construction also reveals how RT-RoPE encodes relative source motion through phase differences between audio tokens. For a horizontal pair $r\in\mathcal I_x$, the relative phase between audio tokens $i$ and $j$ becomes

\begin{equation}
\theta_{hjr}-\theta_{hir}
=
\left(
\theta_{hjr}^{\mathrm{time}}
-
\theta_{hir}^{\mathrm{time}}
\right)
+
\tanh(a_x)\omega_r^x
\left(
\bar{x}_j-\bar{x}_i
\right),
\qquad r\in\mathcal I_x .
\end{equation}

The same formulation applies to the vertical subset $\mathcal I_y$. RT-RoPE therefore augments audio self-attention with signed relative motion displacement.

\section{Experiments}

\subsection{Experimental Settings.} 

\paragraph{Experimental Implementation.}
StereoBind is built upon LTX-2.3, a VA generation model with separate video and audio branches. We incorporate a frozen Motion Track IC-LoRA~\citep{ben2026avcontrol} into the video branch and inject trainable LoRA into the attention and feed-forward of the audio branch, with rank $32$, scaling factor $32$. We employ Qwen-3.5-35B-A3B in the SDS Pipeline and Qwen3-Omni-30B in the VAS Pipeline. StereoBind is trained on our StereoWorld-29K dataset for 8K iterations on 8 NVIDIA H20 GPUs, requiring approximately two days. We adopt the original LTX-2.3 training objective without modification. We use AdamW with a learning rate of $2\times10^{-4}$, $\beta_1=0.9$, and $\beta_2=0.999$, together with a linear learning-rate scheduler. Each training sample contains 121 frames at a resolution of $640\times384$.

\paragraph{Evaluation Protocol.}
We compare StereoBind with representative VA generation baselines, including Ovi~\citep{low2025ovi}, MiniMax-H3, LTX-2.5 and LTX-2.3~\citep{hacohen2026ltx}, as well as video-to-spatial-audio (V2SA) models PrismAudio~\citep{liu2026prismaudio} and See2Sound~\citep{dagli2025see}. Since the VA baselines do not support motion inputs, we augment their prompts with directional motion descriptions to encourage the sound source to follow the desired motion. Since V2SA models do not generate videos, we use StereoBind to produce the input videos, then use the generated videos as inputs to each V2SA model to synthesize their audio outputs. We assess generation performance from three aspects: visual quality using VBench~\citep{huang2024vbench}, audio quality using Audiobox Aesthetics~\citep{tjandra2025meta} and AVGen-Bench~\citep{zhou2026avgen}, and stereo spatial fidelity using our SWBench. Further details of SWBench are provided in Appendix~\ref{sec:appendix_swbench}.

\paragraph{Evaluation Metrics.}
For visual quality, we report Subject Consistency, Motion Smoothness, and Imaging Quality from VBench~\citep{huang2024vbench}. Audio quality is evaluated using audio Production Quality (PQ) from Audiobox Aesthetics~\citep{tjandra2025meta} and AV Sync from Synchformer~\citep{iashin2024synchformer} from AVGen-Bench~\citep{zhou2026avgen}. For stereo spatial fidelity, we first downmix the generated audio to mono and re-spatialize it with our VAS Pipeline to construct the reference stereo audio. We report Interaural Level Difference Wasserstein Distance (ILD-W) for interaural level-difference consistency, Sound Event Localization and Detection Accuracy (SELD-Acc) from SAVGBench~\citep{shimada2026savgbench} for audiovisual spatial alignment, Stereo Magnitude Ratio Error (SMR-Err) for stereo spatial balance, and Spatial-AST Angular Consistency (AST-Ang) and Spatial-AST Calibration (AST-Cal) based on Spatial-AST~\citep{zheng2024bat} for learned stereo consistency. Detailed metrics definitions are provided in Appendix~\ref{sec:appendix_swbench}.

\begin{table*}[t]
\centering
\caption{
\textbf{Comparison results on joint video-audio generation.}  Best results are highlighted in bold, and second-best results are underlined.
}
\label{tab:quantitative_results}

\scriptsize
\setlength{\tabcolsep}{2.8pt}
\renewcommand{\arraystretch}{1.1}

\begin{tabular}{llccc|cc|ccccc}
\toprule
&
& \multicolumn{3}{c|}{{Visual Quality}}
& \multicolumn{2}{c|}{{Audio Quality}}
& \multicolumn{5}{c}{{Stereo Spatial Fidelity}} \\
\cmidrule(lr){3-5}
\cmidrule(lr){6-7}
\cmidrule(lr){8-12}

\textbf{Model Type}
& \textbf{Method}
& \makecell{{Subject}\\{Cons.} $\uparrow$}
& \makecell{{Motion}\\{Smooth.} $\uparrow$}
& \makecell{{Imaging}\\{Qual.} $\uparrow$}
& \makecell{{Audio}\\{PQ} $\uparrow$}
& \makecell{{AV}\\{Sync} $\downarrow$}
& \makecell{{ILD}\\{-W} $\downarrow$}
& \makecell{{SELD}\\{-Acc} $\uparrow$}
& \makecell{{SMR}\\{-Err} $\downarrow$}
& \makecell{{AST}\\{-Ang} $\uparrow$}
& \makecell{{AST}\\{-Cal} $\uparrow$} \\
\midrule

\multirow{3}{*}{\makecell{\textbf{VA}\\\textbf{Models}}}
& Ovi~\citep{low2025ovi}
& \underline{0.962} & 0.993 & 0.681
& 5.70 & \underline{0.278}
& 3.744 & 0.571 & 65.58 & 0.196 & 0.0261 \\

& MiniMax-H3
& \textbf{0.968} & \textbf{0.996} & \underline{0.690}
& \underline{6.81} & 0.301
& 3.435 & \underline{0.725} & \underline{24.64} & 0.257 & \underline{0.131} \\

& LTX-2.5~\citep{hacohen2026ltx}
& 0.950 & \underline{0.995} & 0.655
& 6.70 & 0.367
& 3.565 & 0.571 & 37.62 & 0.237 & 0.122 \\

& LTX-2.3~\citep{hacohen2026ltx}
& 0.953 & \underline{0.995} & 0.653
& 6.68 & 0.374
& 3.574 & 0.617 & 39.34 & 0.221 & 0.119 \\

\midrule

\multirow{2}{*}{\makecell{\textbf{V2SA}\\\textbf{Models}}}
& PrismAudio~\citep{liu2026prismaudio}
& -- & -- & --
& 6.02 & 0.479
& \underline{3.426} & 0.473 & 44.55 & 0.189 & 0.036 \\

& See2Sound~\citep{dagli2025see}
& -- & -- & --
& 5.10 & 0.406
& 6.30 & 0.375 & 30.48 & \underline{0.292} & 0.119 \\

\midrule

\textbf{Ours}
& \textbf{StereoBind}
& 0.955 & \textbf{0.996} & \textbf{0.697}
& \textbf{6.86} & \textbf{0.262}
& \textbf{2.754} & \textbf{0.757} & \textbf{14.86}
& \textbf{0.465} & \textbf{0.314} \\

\bottomrule
\end{tabular}
\end{table*}

\begin{table*}[t]
\centering
\caption{
\textbf{Ablation study of StereoBind on StereoWorldBench (SWBench). }
Best results are highlighted in bold, and second-best results are underlined.
}
\label{tab:ablation}
\scriptsize
\setlength{\tabcolsep}{4pt}
\renewcommand{\arraystretch}{0.9}

\begin{tabular}{lccccc}
\toprule
\textbf{Variant}
& \makecell{{ILD}\\{-W} $\downarrow$}
& \makecell{{SELD}\\{-Acc} $\uparrow$}
& \makecell{{SMR}\\{-Err} $\downarrow$}
& \makecell{{AST}\\{-Ang} $\uparrow$}
& \makecell{{AST}\\{-Cal} $\uparrow$} \\
\midrule

w/o Residual Track RoPE
& 2.800
& 0.653
& 19.28
& 0.308
& 0.219 \\

w/o Spatial Track Encoder
& 3.336
& 0.669
& 27.38
& 0.282
& 0.198 \\

w/o VMB
& 3.079
& 0.506
& 31.16
& 0.256
& 0.216 \\

VMB Tokens ($N_{\mathrm{VMB}}=16$)
& 2.785
& 0.627
& 18.65
& 0.322
& 0.245 \\

VMB Tokens ($N_{\mathrm{VMB}}=32$)
& \textbf{2.751}
& 0.691
& 20.47
& 0.311
& 0.246 \\

VMB Tokens ($N_{\mathrm{VMB}}=128$)
& 2.756
& \underline{0.747}
& \textbf{13.47}
& \textbf{0.471}
& \textbf{0.321} \\

\midrule

\textbf{StereoBind (Full, $N_{VMB}=64$)}
& \underline{2.754}
& \textbf{0.757}
& \underline{14.86}
& \underline{0.465}
& \underline{0.314} \\

\bottomrule
\end{tabular}
\end{table*}

\subsection{Experimental Results.}

\paragraph{Quantitative Results.}

We compare StereoBind with joint VA generation models and V2SA methods in Table~\ref{tab:quantitative_results}. StereoBind achieving the best Motion Smoothness, Imaging Quality, Audio PQ, and AV Sync while maintaining comparable Subject Consistency and audio quality. More importantly, it outperforms all baselines on stereo spatial fidelity metrics, achieving an ILD-W of 2.754, SELD-Acc of 0.757, SMR-Err of 14.86, AST-Ang of 0.465, and AST-Cal of 0.314. These gains demonstrate improved dynamic spatial correspondence without compromising generation quality. Our method also achieves the best human perceptual results, as reported in Appendix~\ref{sec:human_study}.

\paragraph{Qualitative Results.}

\begin{figure*}[t]
  \centering
  \includegraphics[width=\linewidth]{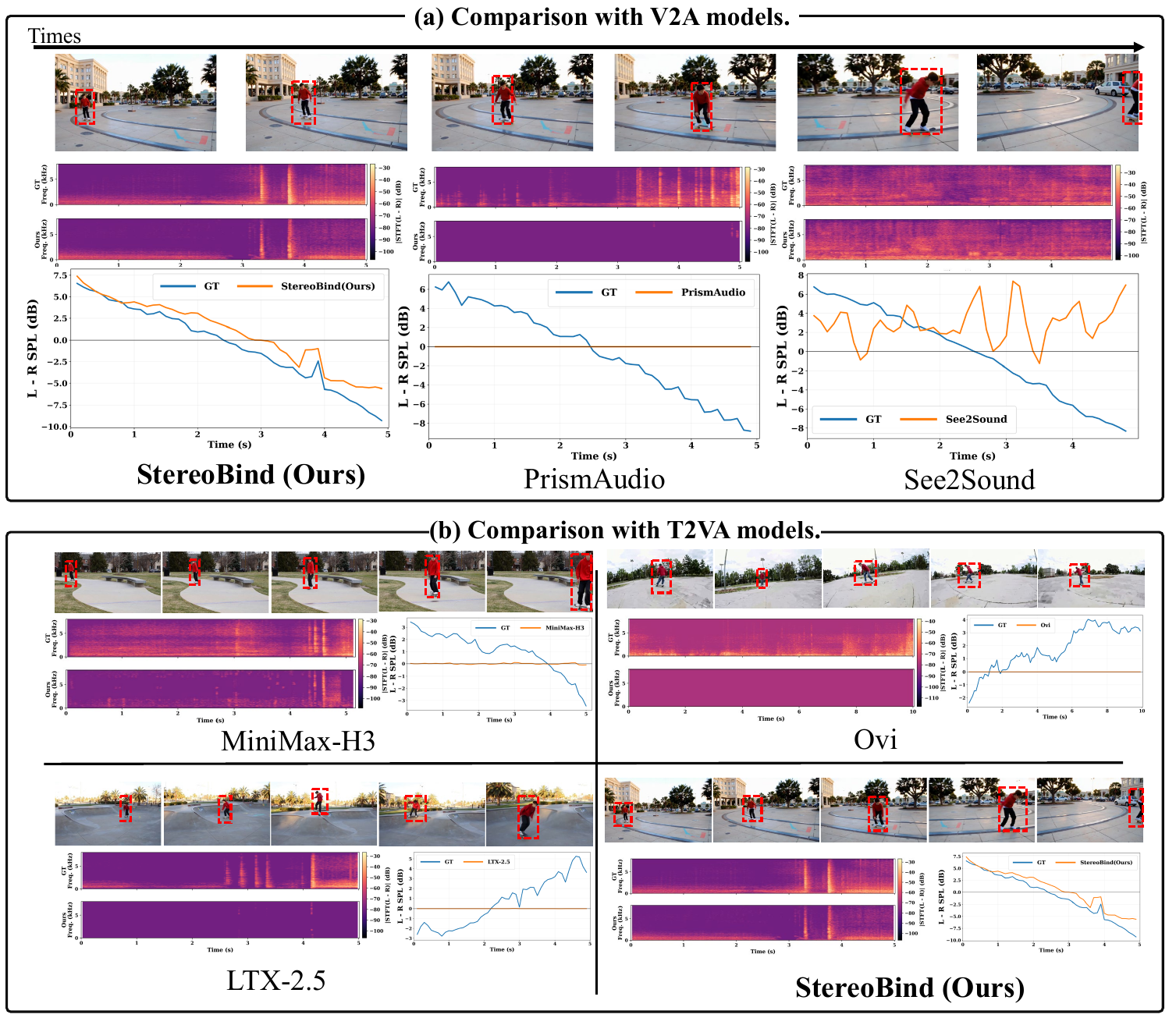}
  \caption{Qualitative experimental results on SWBench. We compare StereoBind with open-source VA models and V2SA models. The sound source is highlighted with a red bounding box.}
  \label{fig:qualitative}
\end{figure*}

We conduct qualitative comparisons using two visualizations, Signed Inter-channel Level Difference (SILD) and Inter-channel Differential Spectrogram (IDS). SILD visualizes the signed inter-channel level difference over time, where positive and negative values indicate left- and right-channel dominance, respectively. IDS visualizes the time-frequency structure of the inter-channel difference using the Short-Time Fourier Transform (STFT). Detailed formulations are provided in Appendix~\ref{sec:appendix_qualitative_vis}. As shown in Fig.~\ref{fig:qualitative}, existing methods exhibit limited dynamic spatial correspondence. Their SILD trajectories often deviate from the reference spatial evolution, while their IDS patterns show inconsistent structures. In contrast, StereoBind produces coherent stereo dynamics that closely follow the source motion, demonstrating stronger dynamic spatial correspondence. For more qualitative results, please refer to Appendix~\ref{sec:appendix_qual} and Supplementary Material.

\subsection{Ablation Study.}

To assess the contribution of key components in StereoBind, we conduct ablation experiments focusing on four factors: (1) RT-RoPE, (2) STE, (3) VMB, and (4) the number of VMB Tokens. All ablation experiments are conducted on SWBench. As summarized in Table~\ref{tab:ablation}, each factor contributes to stereo spatial fidelity, with the full StereoBind achieving the strongest overall performance.

\textbf{The Impact of Residual Track RoPE.}
We evaluate the contribution of RT-RoPE by removing the track-dependent residual phases. As shown in Table~\ref{tab:ablation}, removing RT-RoPE reduces AST-Ang and AST-Cal, indicating degraded learned spatial representations. This suggests that augmenting RoPE with relative source displacement is important for capturing relative spatial dynamics.

\textbf{The Impact of Spatial Track Encoder.}
We remove the STE, which provides absolute spatial conditioning through Track-Adaptive modulation. This variant substantially increases SMR-Err and degrades both AST-Ang and AST-Cal, confirming the importance of absolute source-position information for spatially coherent stereo generation.

\textbf{The Impact of VMB.}
Removing VMB degrades overall spatial fidelity, demonstrating its importance in integrating motion and visual context for audio generation. We further study the capacity of VMB Tokens by varying $N_{\mathrm{VMB}}$. Although the effect is not strictly monotonic across individual metrics, a larger token budget generally provides greater capacity to preserve source-aware visual and motion context. Performance largely saturates at $N_{\mathrm{VMB}}=64$ and further increasing the token number to 128 yields marginal gains while introducing higher memory overhead. Considering the trade-off between spatial fidelity and computational cost, we adopt $N_{\mathrm{VMB}}=64$ in the final model.

\section{Conclusion}
We introduce StereoBind, a framework that extends stereo video-audio generation from semantic and temporal correspondence to dynamic spatial correspondence. StereoBind combines VMB for cross-modal motion conditioning, STE for absolute spatial conditioning, and RT-RoPE for relative motion modeling. We further construct StereoWorld-29K for training and StereoWorldBench for evaluating dynamic spatial correspondence. Experiments show that StereoBind improves spatial consistency with visual source motion while preserving overall audiovisual generation quality. Future work will extend dynamic spatial correspondence to richer 3D scenes, multiple sound sources, and more complex source--listener dynamics.

\bibliographystyle{splncs04}
\bibliography{main}

\newpage

\appendix

\section{Detailed Architecture of the Spatial Track Encoder}

The Spatial Track Encoder (STE) converts a frame-wise 2D motion track into temporally aligned spatial conditioning for the audio branch. Given a motion track
$P\in\mathbb{R}^{T_p\times2}$, each coordinate is first projected into a latent motion space $H_0=W_{\mathrm{in}}P,$ where $W_{\mathrm{in}}$ denotes the input projection layer. In our implementation, the track sequence contains $T_p=121$ temporal positions, and the latent dimension is set to $d_s=256$. To capture temporal dependencies of the source motion, the projected track features are processed by a temporal Transformer encoder followed by LayerNorm:

\begin{equation}
H=
\operatorname{LN}
\left(
\operatorname{TemporalTransformer}(H_0)
\right).
\end{equation}

The resulting representation contains frame-level motion context. Since the audio branch operates on a different temporal resolution, we further resample the track representation to the audio token sequence length $\widetilde{H}=\mathcal{R}_{N_a}(H)$, where $\mathcal{R}_{N_a}(\cdot)$ denotes linear temporal interpolation and $N_a$ is the number of audio tokens. The shared representation $\widetilde{H}$ is computed once during each forward pass and provides temporally aligned geometric information for subsequent Transformer blocks. To adapt the shared spatial representation to different Transformer layers, we introduce a layer-specific Track-Adaptive modulation mechanism. For each Transformer block $\ell$, a learnable adaptive table
$B_\ell\in\mathbb{R}^{1\times3D_a}$
is added to the projected track features:

\begin{equation}
[\beta_\ell,\gamma_\ell,g_\ell]
=
\mathcal{U}_3
\left(
\operatorname{Linear}_u
(
\operatorname{SiLU}(\widetilde{H})
)
+
B_\ell
\right),
\end{equation}

where $\mathcal{U}_3(\cdot)$ splits the feature dimension into three equal groups. The resulting parameters
$\beta_\ell,\gamma_\ell,g_\ell
\in
\mathbb{R}^{N_a\times D_a}$
represent the layer-specific spatial modulation signals. Specifically, $\beta_\ell$ and $\gamma_\ell$ modulate the audio queries during VMB retrieval, enabling the audio branch to access motion-aware visual context, while $g_\ell$ controls the strength of the retrieved spatial update. Through this design, STE provides explicit geometric conditioning while allowing each Transformer layer to dynamically adapt the shared motion representation to its evolving audio feature space.

\section{Additional Details of Visual-Audio Spatialization Pipeline}
\label{sec:appendix_vas}

The VAS Pipeline converts an existing audio-video clip into a spatially supervised stereo example. The complete pipeline consists of four stages: sound-source identification, candidate grounding and audiovisual verification, motion-track extraction, and geometry-aware stereo rendering.

\subsection{Visual-Audio Source Discovery}
Given an input video and its original audio, we first extract a 16-kHz mono waveform and jointly provide the video and audio to Qwen3-Omni-30B-A3B-Instruct. The model is instructed to identify the dominant audible event and associate it with a concrete visible physical source, outputting a short grounding phrase, such as guitar, dog, person, for Grounded-SAM localization. The core instruction used for sound-source identification is shown in Figure~\ref{fig:prompt}.

\subsection{Souding Source Verification}

The generated grounding phrase is passed to GroundingDINO and SAM2 to obtain candidate bounding boxes and segmentation masks. We preserve multiple plausible instances whenever the same semantic category appears more than once. For each candidate, we construct a verification view in which the candidate remains visible, preserving contextual information while clearly indicating the entity to be assessed. We then perform candidate-wise audiovisual verification using Qwen3-Omni-30B-A3B-Instruct. Each candidate video is paired with the original mono audio, and the model evaluates whether the entity produces the dominant sound. The verifier explicitly evaluates four complementary aspects: semantic consistency, action--acoustic consistency, temporal synchronization, and physical causal plausibility. The corresponding scores are aggregated through a weighted average to obtain an overall confidence score, and the candidate with the highest confidence is selected as the final sound source. The core verification instruction is shown in Figure~\ref{fig:prompt}.

\subsection{Sound Source Tracking}

After selecting the sound source, we apply AllTracker~\citep{harley2025alltracker} to its verified mask sequence to obtain a sparse motion track, which is further smoothed using Gaussian filtering for the horizontal coordinates and third-order polynomial fitting for the vertical coordinates before being used as spatial supervision. This produces a normalized $T_p\times2$ motion track that is subsequently used both for spatial-audio construction and as motion conditioning during model training.

\subsection{VGGT-Based Dynamic Stereo Rendering}

The 2D motion track does not directly provide physical source-listener geometry. We therefore use VGGT~\citep{wang2025vggt} to recover scene geometry and camera parameters. Let $(u_t,v_t)$ denote the image position of the tracked source and $z_t$ its estimated depth. Given the camera intrinsic matrix $\mathbf{K}_t$, the source is back-projected to camera coordinates as

\begin{equation}
\mathbf{p}^{\mathrm{cam}}_t
=z_t
\mathbf{K}^{-1}_t
\begin{bmatrix}
u_t\
v_t\
1
\end{bmatrix}.
\end{equation}

Since monocular geometry contain a global scale ambiguity, we use a median-distance anchoring strategy to map the recovered geometry to a physically plausible metric scale. The default anchor places the median source distance at $1.5$~m. The reconstructed camera is treated as a virtual listener. Two virtual listener are placed symmetrically around the listener center along the horizontal listener axis. This converts the source-camera geometry into two time-varying propagation paths. For source position $\mathbf{s}_t$, listener center $\mathbf{r}_t$, normalized left-to-right direction $\mathbf{d}_t$, and interaural distance $d_e$, the virtual receiver locations are

\begin{equation}
\begin{aligned}
\mathbf{r}^{L}_t
&=
\mathbf{r}_t
-
\frac{d_e}{2}\mathbf{d}_t,
\\
\mathbf{r}^{R}_t
&=
\mathbf{r}_t
+
\frac{d_e}{2}\mathbf{d}_t.
\end{aligned}
\end{equation}

The expected direct-path interaural delay is therefore

\begin{equation}
\Delta \tau_t
=
\frac{
\left\lVert \mathbf{s}_t-\mathbf{r}^{R}_t \right\rVert_2
-
\left\lVert \mathbf{s}_t-\mathbf{r}^{L}_t \right\rVert_2
}{c}.
\end{equation}

where $c=343$~m/s is the speed of sound. We then use gpuRIR~\citep{diaz2021gpurir} to simulate the room impulse response from the moving source to both virtual receivers. The base scene configuration uses a $10\times10\times4$~m room with $T_{60}=0.25$~s. We use an 8~ms early-response window, an 8~ms transition interval, a $0$-dB early gain, and a $-6$-dB late-reverberation gain. Given mono waveform $a(\tau)$ and time-varying left/right impulse responses $h_t^L$ and $h_t^R$, stereo rendering can be written conceptually as

\begin{equation}
a_t^L = a * h_t^L,
\qquad
a_t^R = a * h_t^R,
\end{equation}

These operations allows ITD, level differences, and reverberant cues to evolve consistently with the reconstructed visual-source track.

\begin{table}[t]
\centering
\small
\caption{Sampling weights for dynamic motion families.}
\label{tab:motion_distribution}
\begin{tabular}{lc}
\toprule
Motion family & Weight \\
\midrule
Horizontal left-to-right & 0.20 \\
Horizontal right-to-left & 0.20 \\
Diagonal left-to-right & 0.12 \\
Diagonal right-to-left & 0.12 \\
Curved left-to-right & 0.12 \\
Curved right-to-left & 0.12 \\
Vertical upward & 0.04 \\
Vertical downward & 0.04 \\
Stop-and-go left-to-right & 0.02 \\
Stop-and-go right-to-left & 0.02 \\
\bottomrule
\end{tabular}
\end{table}

\begin{table}[t]
\centering
\small
\caption{Semantic-family groups used for different spatial configurations.}
\label{tab:semantic_family}
\resizebox{\linewidth}{!}{
\begin{tabular}{lll}
\toprule
Motion & Extent & Semantic families \\
\midrule
Static &
Point-like &
stationary animal, small household device, public device, small mechanical device, tonal object \\

Static &
Area-like &
large household appliance, water source, large mechanical machine, airflow machine, fire/flame source \\

Dynamic &
Point-like &
moving animal, wheeled object, compact vehicle, mobile robot, small mobile machine \\

Dynamic &
Area-like &
large vehicle, floor-cleaning machine, outdoor machine, industrial mobile machine \\
\bottomrule
\end{tabular}
}
\end{table}

\section{Additional Details of Spatial Data Synthesis Pipeline}
\label{sec:appendix_sds}

The SDS Pipeline follows a two-stage design: the first stage samples a diverse structured scene specification, while the second stage converts the specification into a generation prompt while preserving the sampled constraints.

\subsection{Spatial Configuration Space}

For the object-centric branch, each sample is first assigned a structured spatial configuration consisting of $\mathcal{C} = \{m, f, e, s\}$, where $m$ denotes the motion type, $f$ the motion family, $e$ the source extent, and $s$ the semantic family. Details are shown in Table~\ref{tab:semantic_family}. The default proportions for static-left and static-right sources are both $0.15$, leaving $0.70$ of samples for dynamic motion. Source extent is sampled as point-like with probability $0.65$ and area-like with probability $0.35$. For dynamic examples, the motion-family distribution is shown in Table~\ref{tab:motion_distribution}. The semantic family is conditioned jointly on motion type and source extent. Based on the sampled configuration, we instantiate a concrete motion track that satisfies the selected motion type and motion family, while the source extent and semantic family constrain the physical identity and spatial characteristics of the sound-producing object. Each sampled motion track is provided to Qwen3.5-35B-A3B, which converts the track into an explicit textual motion description. The generated description is subsequently incorporated into the VA-generation prompt to maintain consistency between textual conditioning and motion-track conditioning.

\subsection{Scene Distribution Bank}

A purely independent sampling strategy tends to repeatedly generate common categories and scenes. We therefore maintain a Scene Distribution Bank (SDB) that records statistics over accepted synthetic samples. SDB maintains online counts for source and scene category, semantic family, motion family and sound events. When selecting among semantic families, underrepresented categories receive larger sampling probability. For a candidate category $c$ with current count $n_c$, its unnormalized sampling weight is

\begin{equation}
w(c)=\frac{1}
{\left(1+n_c\right)^{1.25}}.
\end{equation}

This mechanism favors underrepresented categories while retaining stochasticity. In addition to global frequency statistics, the most recent 5 accepted samples are passed to the Stage~1 as negative diversity context. The model is explicitly instructed to avoid repeating their sound source events and scene configurations. 

\subsection{Object-Centric Scene Generation}

\begin{figure*}[t]
  \centering
  \includegraphics[width=\linewidth]{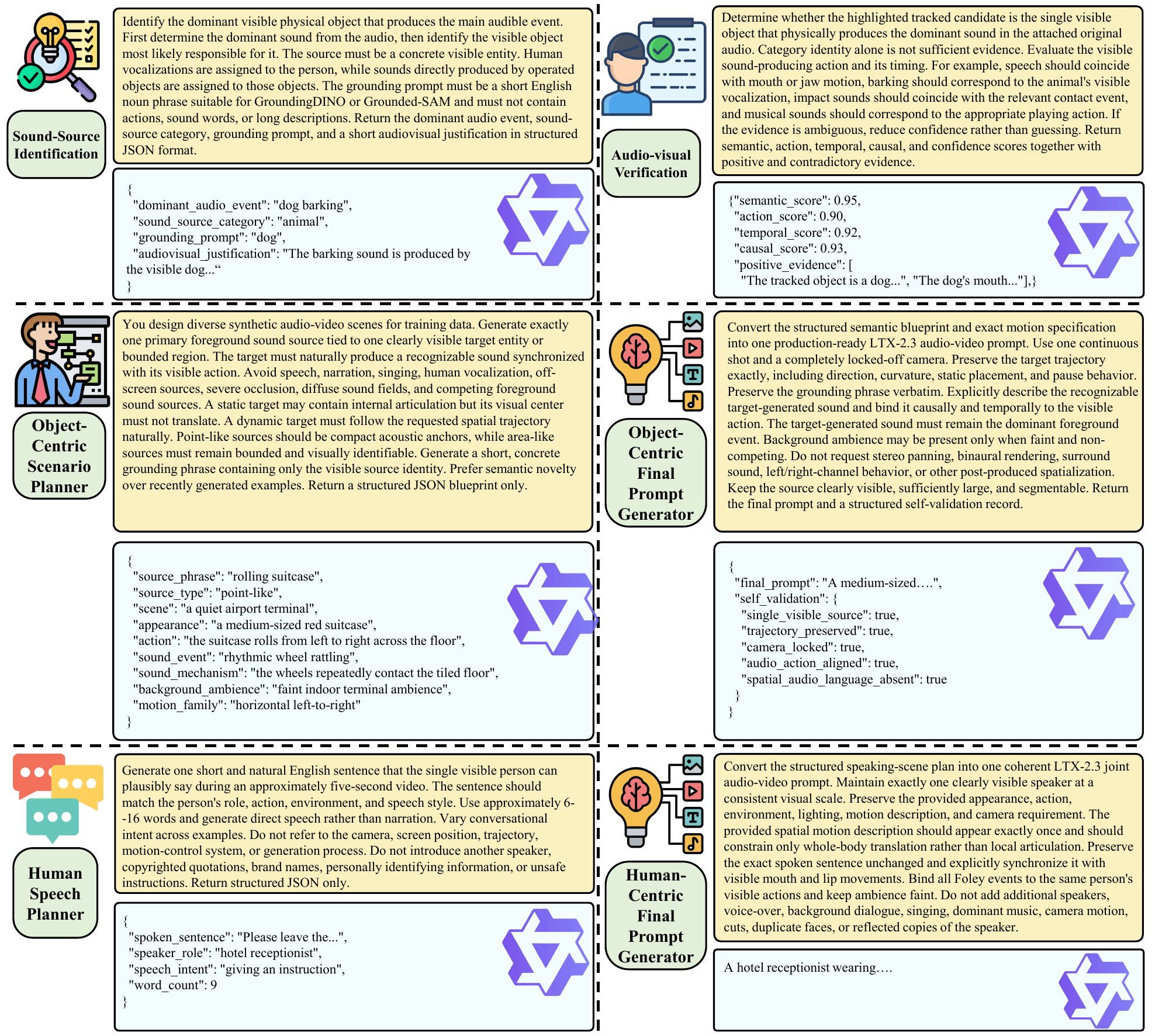}
  \caption{Instructions and output formats of the Qwen-based modules used in StereoWorld-29K construction.}
\label{fig:prompt}
\end{figure*}

For object-centric scenes, Stage~1 utilizes a uses Qwen3.5-35B-A3B to generate a structured semantic blueprint. The model receives the sampled motion type, motion family, motion description, source extent, semantic family, SDB statistics, and recent examples to avoid. The required blueprint contains the source phrase, source category, scene, scene category, visual appearance, source action, recognizable sound events, physical sound-production mechanism, background ambience, acoustic extent, and a short explanation of why the target is visually segmentable. Stage~2 receives the validated blueprint and then converts them into an LTX-2.3-compatible VA-generation prompt. The core instruction is shown in Figure~\ref{fig:prompt}

\subsection{Human-Centric Scene Generation}

Since speech generation introduces additional constraints on speech content and speaker identity, we use a dedicated synthesis branch with two-stage procedure similar to object-centric scene generation for human-speaking scenes. We construct a predefined catalog containing 47 human-speaking profiles and 41 concrete environments by GPT-5.6. Profiles describe plausible speaker identities, actions, sound-producing interactions, and scene types. Environments cover diverse categories including offices, schools, university interiors, etc.. Moving human sources use simple horizontal motion to improve subject stability and facial consistency. 

In Stage~1, the Qwen3.5 serve as a speech planner to generates a fresh spoken sentence conditioned on the sampled speaker role, environment. Spoken content is constrained to approximately 15 English words so that it can be naturally produced within a five-second clip. Stage~2 converts the complete scene specification and generated speech content into an LTX-2.3 VA prompt. The final prompt must contain the exact spoken sentence unchanged, explicitly associate the voice with the speaker.

\section{Distribution Analysis of StereoWorld-29K}
\label{sec:appendix_stereoworld}

\begin{figure*}[t]
  \centering
  \includegraphics[width=\linewidth]{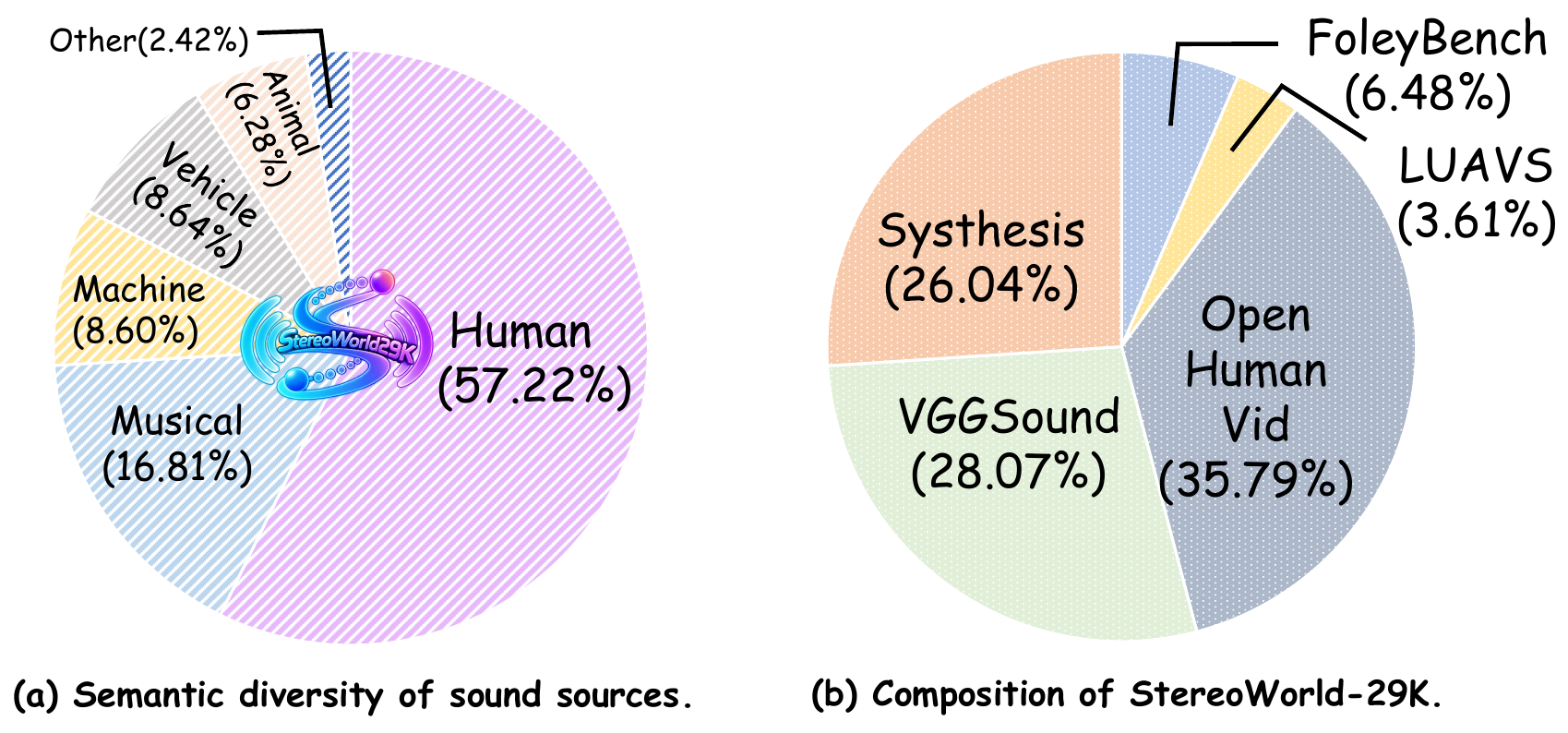}
  \caption{Dataset statistics of StereoWorld-29K. (a) Distribution of semantic categories of visible sound-producing entities. (b) Composition of real-world and synthetic data sources.}
\label{fig:stat1}
\end{figure*}

\begin{figure*}[t]
  \centering
  \includegraphics[width=\linewidth]{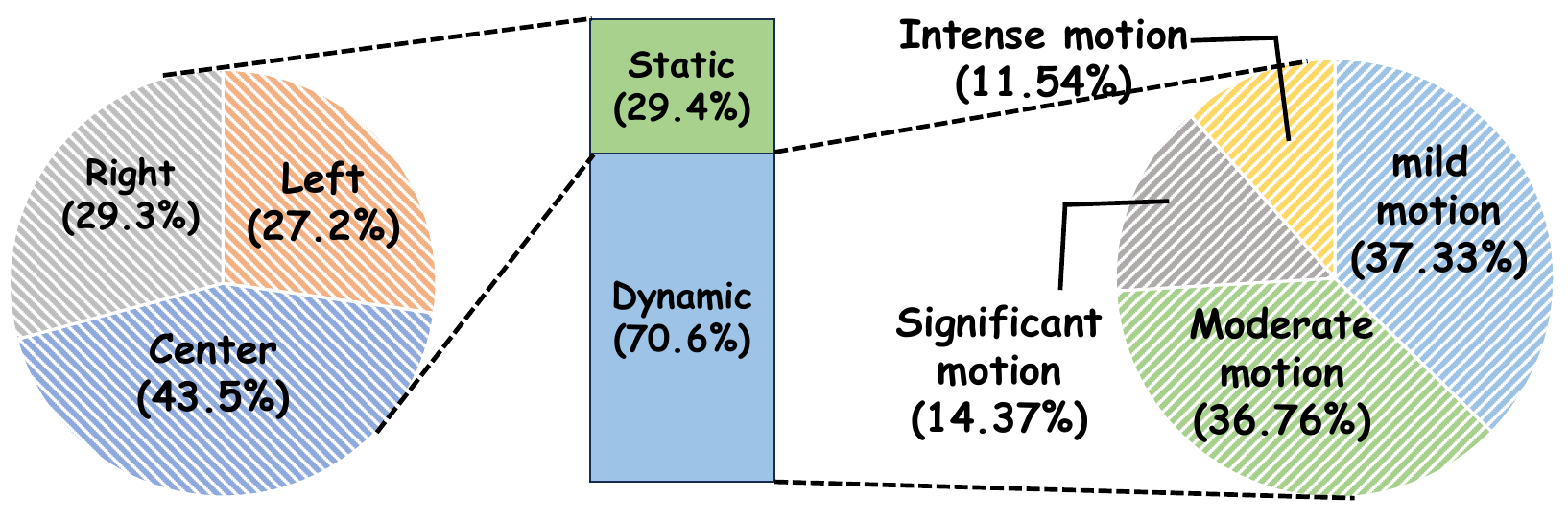}
  \caption{Motion and spatial statistics of StereoWorld-29K. Dynamic samples are categorized into four motion levels according to horizontal source displacement, while static samples are grouped by left, center, and right source locations.}
\label{fig:stat2}
\end{figure*}

To characterize the diversity of StereoWorld-29K, we analyze the dataset from three complementary perspectives: data sources, semantic categories, and motion characteristics. StereoWorld-29K is constructed from multiple real-world and synthetic data sources, providing broad coverage across diverse audio-visual scenarios, which is demonstrated in Figure~\ref{fig:stat1}(b). Additionally, as shown in Figure~\ref{fig:stat1}(a), its semantic distribution further spans a wide range of visible sound-producing entities, demonstrating substantial semantic diversity. For motion analysis, we observe that stereo spatial perception is primarily reflected along the horizontal axis, where left-right movements induce more salient inter-channel variations. Therefore, we divide each video frame horizontally into five equal-width regions and categorize motion intensity according to the number of regions traversed by the sound source. Specifically, as shown in Figure~\ref{fig:stat2}, motion crossing one region are categorized as \textbf{mild motion}, two regions as \textbf{moderate motion}, three regions as \textbf{significant motion}, and four regions as \textbf{intense motion}. These motion levels correspond to different degrees of stereo audio variation, where larger horizontal displacement generally leads to more pronounced temporal changes in stereo cues. For static samples, sound sources are distributed across the left, center, and right portions of the frame, providing broad spatial coverage of different source locations.

\section{Human Perceptual Evaluation}
\label{sec:human_study}

To complement the quantitative evaluation, we conduct a human perceptual study to assess whether the proposed dynamic spatial correspondence translates into a perceptible improvement in immersive audiovisual experience. In particular, this study evaluates whether the generated stereo audio remain consistent with the corresponding visible sound source over time. Specifically, We compare StereoBind with VA generation models Ovi~\citep{low2025ovi}, MiniMax-H3, and LTX-2.5~\citep{hacohen2026ltx}, as well as V2SA models PrismAudio~\citep{liu2026prismaudio} and See2Sound~\citep{dagli2025see}. We recruit 25 volunteers and randomly sample 15 comparison groups from SWBench. Each comparison group contains outputs generated from the same input condition by all evaluated methods, allowing participants to directly compare their perceptual differences. The presentation order of the methods is randomized to reduce ordering bias, and all participants are instructed to wear headphones during evaluation. Participants rate each sample using a five-point Likert scale according to its overall \emph{immersion}, jointly considering (1) audiovisual correspondence, (2) whether the perceived motion naturally follow the visible source, (3) the realism of the stereo spatial effect, and (4) the overall sense of presence. The rating criteria are defined as follows:

\begin{itemize}
    \item \textbf{1 -- Very Poor:} Almost no sense of immersion. Audio and visual content are clearly inconsistent, with unnatural spatial positioning or motion and a strongly disconnected overall experience.
    
    \item \textbf{2 -- Poor:} Weak immersion. Some audiovisual correspondence can be perceived, but noticeable inconsistencies, unnatural spatial behavior, or insufficient spatial perception remain.
    
    \item \textbf{3 -- Fair:} Basic immersion. Audio and visual content are generally consistent, but spatial realism, motion naturalness, or the sense of presence still have clear room for improvement.
    
    \item \textbf{4 -- Good:} Strong immersion. Audio is well aligned with the visual content, spatial position and motion are natural, and the stereo experience is realistic, with only minor perceptual imperfections.
    
    \item \textbf{5 -- Excellent:} Very strong immersion. Audio and visual content are highly coherent, sound position and motion evolve naturally with the visible source, and the stereo rendering provides a convincing sense of spatial presence and immersion.
\end{itemize}

Table~\ref{tab:user_study} reports the mean perceptual scores, while StereoBind achieves the highest mean score. These results indicate that explicitly modeling dynamic spatial correspondence leads to a stronger perceived consistency between visible sound source and stereo audio, resulting in a more immersive audiovisual experience.

\begin{table}[t]
\centering
\caption{
\textbf{Human perceptual evaluation on SWBench.}
Participants rate the overall immersive experience on a five-point Likert scale, considering audiovisual correspondence, spatial motion consistency, stereo realism, and sense of presence. Higher is better.
}
\label{tab:user_study}
\small
\setlength{\tabcolsep}{8pt}
\renewcommand{\arraystretch}{1.1}

\begin{tabular}{lc}
\toprule
\textbf{Method} & \textbf{Immersion Score $\uparrow$} \\
\midrule
Ovi~\citep{low2025ovi}             & 1.2 \\
MiniMax-H3                         & 2.1 \\
LTX-2.5~\citep{hacohen2026ltx}    & \underline{2.3} \\
PrismAudio~\citep{liu2026prismaudio} & 1.5 \\
See2Sound~\citep{dagli2025see}    & 1.1 \\
\midrule
StereoBind (Ours)                  & \textbf{4.2} \\
\bottomrule
\end{tabular}
\end{table}

\section{Additional Qualitative Results}
\label{sec:appendix_qual}

\begin{figure*}[t]
  \centering
  \includegraphics[width=\linewidth]{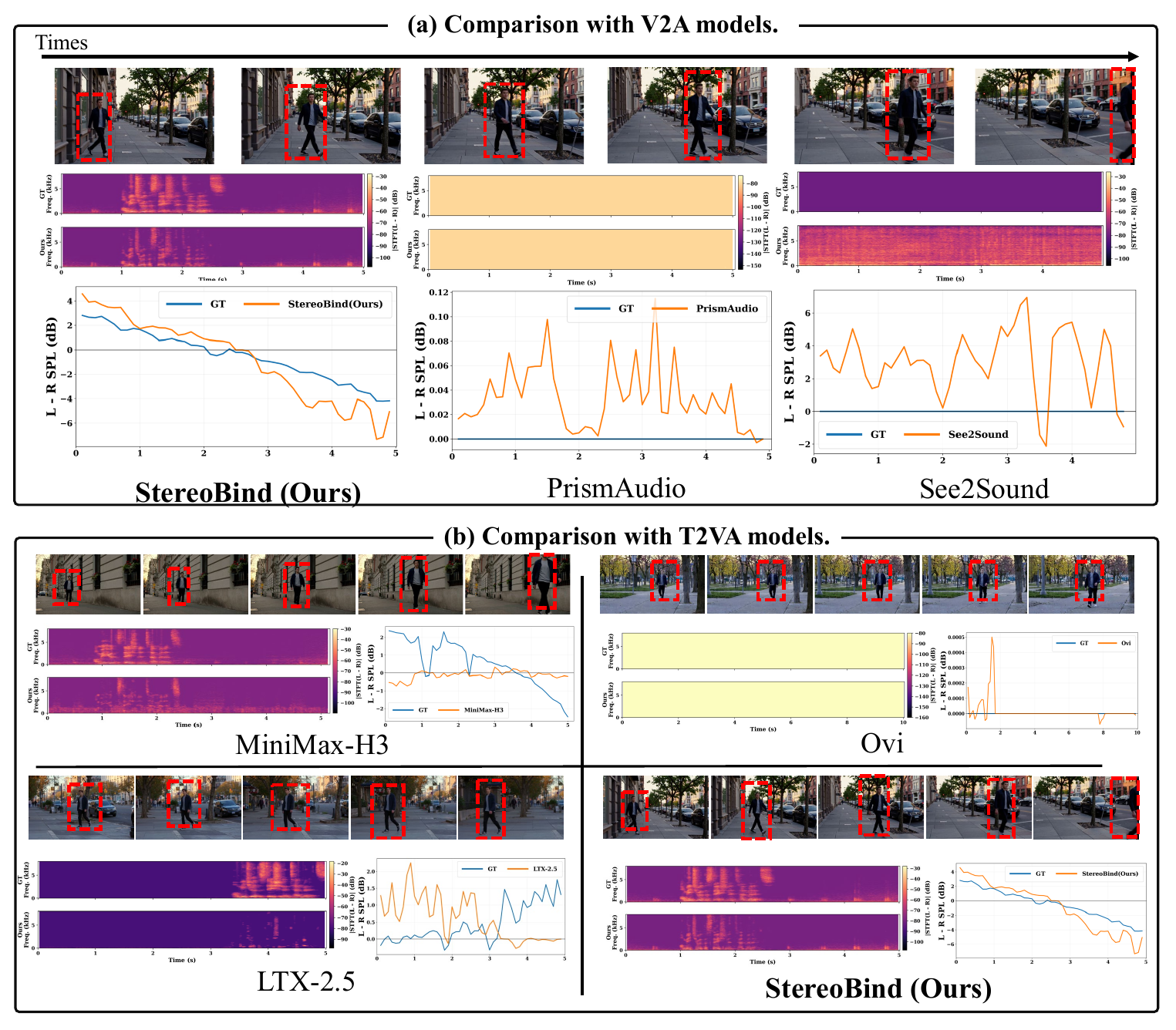}
  \caption{Qualitative experimental results on SWBench. We compare StereoBind with open-source VA models and V2SA models. The sound source is highlighted with a red bounding box.}
  \label{fig:qual1}
\end{figure*}

\begin{figure*}[t]
  \centering
  \includegraphics[width=\linewidth]{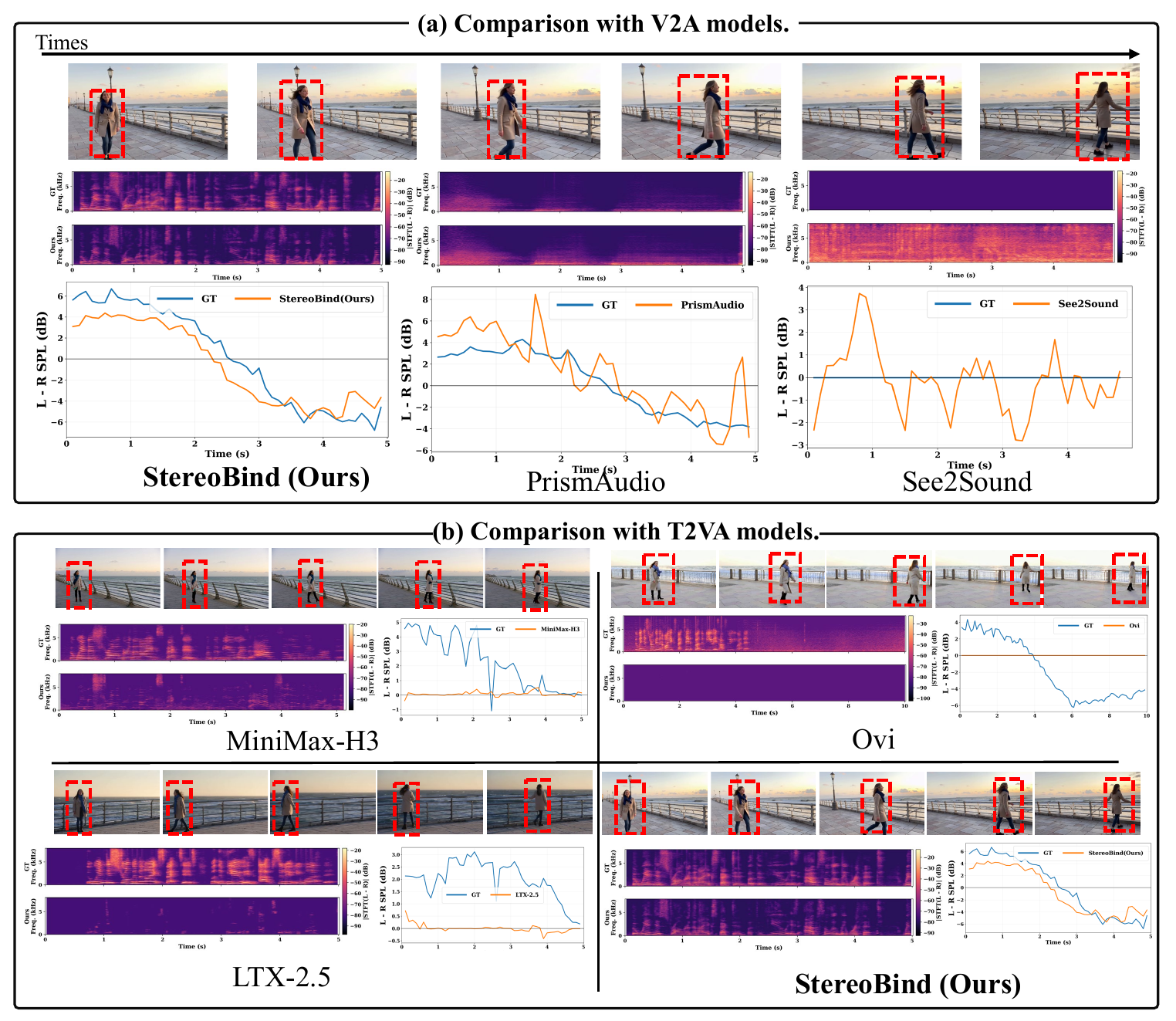}
  \caption{Qualitative experimental results on SWBench. We compare StereoBind with open-source VA models and V2SA models. The sound source is highlighted with a red bounding box.}
  \label{fig:qual2}
\end{figure*}

In this section, we provide additional qualitative comparisons for stereo VA generation in Figure~\ref{fig:qual1} and Figure~\ref{fig:qual2}. The results demonstrate that our method generates spatially coherent stereo audio that consistently follows the motion of visible sound sources. Compared with existing approaches, StereoBind better preserves audio-visual correspondence and produces more temporally consistent spatial transitions.  It is worth noting that some spectrograms in Figure~\ref{fig:qual1} appear as nearly uniform color regions. This is because the generated audio in these cases corresponds to stationary white noise, whose spectral energy remains approximately constant over time, resulting in visually homogeneous spectrogram representations.

\section{Details of Qualitative Visualization}
\label{sec:appendix_qualitative_vis}

To provide an intuitive analysis of dynamic spatial correspondence, we complement the quantitative evaluation with two audio-side visualizations: \textbf{Signed Inter-channel Level Difference (SILD)} and \textbf{Inter-channel Differential Spectrogram (IDS)}. The former characterizes the temporal evolution of left--right dominance, while the latter reveals the time--frequency structure of inter-channel discrepancies. For all compared methods, the stereo audio are first resampled to a common sampling rate of $f_s=16\,\mathrm{kHz}$ and truncated to the same duration.

\subsection{Signed Inter-channel Level Difference}
\label{sec:appendix_sild}

\paragraph{Computation.}

Let $x_L[n]$ and $x_R[n]$ denote the left- and right-channel waveforms, respectively. We first compute short-time root-mean-square (RMS) amplitudes independently for the two channels. For channel $c\in\{L,R\}$ and temporal window $t$, the RMS amplitude is

\begin{equation}
A_c(t)
=
\sqrt{
\frac{1}{N_w}
\sum_{n=0}^{N_w-1}
x_c^2[tH+n]
+
\epsilon
},
\end{equation}

where $N_w$ denotes the window length, $H$ is the temporal hop size, and $\epsilon$ is a small constant for numerical stability. In our implementation, we use a $200\,\mathrm{ms}$ analysis window and a $100\,\mathrm{ms}$ hop. The RMS amplitudes are then converted into the logarithmic domain to represent the relative channel energy level:

\begin{equation}
L_c(t)
=
20\log_{10}
\left(
A_c(t)+\epsilon
\right),
\end{equation}

where $L_c(t)$ denotes the short-time logarithmic RMS level of channel $c$. The Signed Inter-channel Level Difference (SILD) is computed as the difference between the two channel levels:

\begin{equation}
\mathrm{SILD}(t)
=
L_L(t)-L_R(t).
\end{equation}

Positive and negative values indicate stronger acoustic energy in the left and right channels, respectively, providing an intuitive representation of the temporal evolution of stereo directionality.

\paragraph{Interpretation.}
The sign of SILD directly represents instantaneous stereo dominance. A positive value, $\mathrm{SILD}(t)>0$, indicates that the left channel contains stronger acoustic energy, whereas a negative value, $\mathrm{SILD}(t)<0$, indicates stronger energy in the right channel. Values around zero correspond to approximately balanced left--right energy. This signed representation is particularly suitable for analyzing moving sound sources. For example, when a visible source moves from the left side of the scene toward the right, a spatially consistent stereo signal is expected to exhibit a corresponding transition from positive toward negative SILD values. Conversely, a nearly constant SILD track may indicate insufficient temporal spatialization. SILD therefore converts the stereo signal into an intuitive one-dimensional temporal track that can be directly compared with the horizontal motion of the visible source. In our qualitative figures, we plot the ground truth and different generation methods on the same temporal axis, allowing differences in stereo directionality and temporal transitions to be observed directly.

\subsection{Inter-channel Differential Spectrogram}
\label{sec:appendix_ids}

\paragraph{Computation.}
While SILD summarizes the stereo relation into a broadband level difference, it does not reveal how inter-channel discrepancies are distributed across time and frequency. We therefore additionally visualize the \textbf{Inter-channel Differential Spectrogram (IDS)}. Given the stereo waveform, we first construct the differential signal $d[n]=x_L[n]-x_R[n].$ This operation suppresses signal components that are identical in the two channels and emphasizes components that differ between them. The constant factor does not affect the qualitative time--frequency structure considered here. We then apply the Short-Time Fourier Transform (STFT) to the differential signal:

\begin{equation}
D(t,k)
=
\sum_{n=0}^{N_{\mathrm{FFT}}-1}
d[tH+n]\,
w[n]\,
\exp
\left(
-j\frac{2\pi kn}{N_{\mathrm{FFT}}}
\right),
\end{equation}

where $w[n]$ is the analysis window, $N_{\mathrm{FFT}}$ is the FFT size, and $H$ denotes the hop size. The implementation uses a Hann window with $N_{\mathrm{FFT}}=1024$ and $H=256$ samples. At $16\,\mathrm{kHz}$ sampling rate, these correspond to approximately $64\,\mathrm{ms}$ analysis windows and $16\,\mathrm{ms}$ temporal hops. The displayed IDS is the logarithmic magnitude of this differential STFT:

\begin{equation}
\mathrm{IDS}(t,k)
=
20\log_{10}
\left[
\max
\left(
|D(t,k)|,
10^{F_{\mathrm{floor}}/20}
\right)
\right],
\end{equation}

where $F_{\mathrm{floor}}=-80\,\mathrm{dB}$ is used to suppress extremely weak components. Importantly, all methods within the same comparison use a shared color scale. The upper bound is determined by the maximum differential spectral magnitude among all compared signals, while the displayed dynamic range is shared across methods. Consequently, stronger or weaker inter-channel structures cannot be artificially amplified through independent per-method normalization.

\paragraph{Interpretation.}
IDS visualizes where and when the two stereo channels differ in the time--frequency domain. If the generated signal approaches duplicated mono audio, i.e., $x_L[n]\approx x_R[n]$, then $d[n]$ becomes small and the differential spectrogram contains little energy. In contrast, pronounced stereo differences produce stronger structures in the IDS. The visualization also reveals whether stereo discrepancies evolve continuously over time. Unlike SILD, however, IDS uses the magnitude $|D(t,k)|$ and therefore does not preserve the directional sign of the stereo difference. A strong IDS response indicates a substantial inter-channel discrepancy but does not determine whether the acoustic energy is biased toward the left or right channel. For qualitative presentation, uniformly sampled video keyframes are placed above the IDS maps. This makes it possible to visually compare changes in source position with the corresponding evolution of stereo difference patterns.

\section{Details of StereoWorldBench}
\label{sec:appendix_swbench}

\subsection{StereoWorldBench Sample Composition}

StereoWorldBench contains 95 evaluation samples with each sample consists of a motion track paired with a corresponding textual prompt. The benchmark is constructed independently from the training set. Specifically, GPT-5.6 is used to generate both the scene prompt and its associated sparse motion track, then the generated sparse track is subsequently interpolated to a dense track of size $121\times2$. No prompt--track pair in StereoWorldBench is reused from StereoWorld-29K.

\paragraph{Spatial and Motion Distribution.}
Among the 95 samples, 37 are static and 58 are dynamic. The static subset contains 15 left-positioned and 22 right-positioned sources. Their normalized horizontal coordinates occupy clearly separated regions: left-side sources lie within $x\in[0.08,0.35]$, with a mean position of 0.197, whereas right-side sources lie within $x\in[0.64,0.92]$, with a mean of 0.785. All dynamic samples are designed to exhibit intense source motion. Among them, 27 follow left-to-right motion, 22 follow right-to-left motion, and the remaining 9 contain other non-static track. Across all dynamic samples, the track collectively cover approximately $x\in[0.059,0.939]$ of the normalized image width.

\paragraph{Semantic Composition.}
StereoWorldBench covers diverse visible sound-producing entities while maintaining a clear dominant audiovisual source in each sample. Human subjects constitute the largest category, with 74 samples (77.89\%), including pedestrians, workers, performers, and other interacting people. Machine or object sources account for 12 samples (12.63\%), including drones, vehicles, appliances, and robots, while animal sources contribute the remaining 9 samples (9.47\%).

\paragraph{Acoustic Composition.}
The benchmark also spans multiple types of sound-producing events. Speech, dialogue, and broadcast audio form the largest group with 47 samples (49.47\%). Motion- and contact-related sounds account for 16 samples (16.84\%), while mechanical, electronic, or stationary-object sounds account for 13 samples (13.68\%). Musical or non-speech vocal performances and animal-related sounds each contribute 9 samples (9.47\%), with one additional sample containing an artificial signal sound. This distribution provides a broad range of acoustic structures while retaining explicit correspondence between the visible source and its emitted sound.

\subsection{Spatial-Audio Evaluation Metrics}

We evaluate generated stereo audio against its reference using complementary criteria. The generated audio are resampled to $16\,\mathrm{kHz}$, trimmed to their common duration, and divided into $100\,\mathrm{ms}$ windows with a $100\,\mathrm{ms}$ hop.

\paragraph{Interaural Level Difference Wasserstein Distance (ILD-W).}

ILD-W measures the consistency of interaural level cues between the generated and reference stereo audio. Given the short-time Fourier spectra of the left and right channels, denoted by $A_{L,t}(f)$ and $A_{R,t}(f)$ at window $t$, the interaural level difference is defined as

\begin{equation}
\mathrm{ILD}_{t}(f)
=
20\log_{10}
\left(
\frac{|A_{L,t}(f)|+\epsilon}
{|A_{R,t}(f)|+\epsilon}
\right),
\end{equation}

where a positive value indicates stronger energy in the left channel, while a negative value indicates stronger energy in the right channel. Following the high-frequency regime in which interaural level differences provide informative spatial cues, we compute ILD over the $1.7$--$4.6\,\mathrm{kHz}$ frequency band. For each window, we apply a Hann window before Fourier analysis and use the combined binaural magnitude $w_t(f)=|A_{L,t}(f)|+|A_{R,t}(f)|$ to weight individual frequency bins. We discard low-energy components, clip the remaining ILD values to $[-24,24]\,\mathrm{dB}$, and construct a normalized weighted histogram with $K=400$ bins. The resulting histogram $H_t$ represents the distribution of interaural level differences within each temporal window. The corresponding cumulative distribution function is obtained by cumulatively summing the normalized histogram probabilities:
\begin{equation}
F_t(x)
=
\sum_{k:\,c_k \leq x}
H_t(k),
\end{equation}
where $c_k$ denotes the center of the $k$-th ILD bin. Applying this definition to the reference and generated histograms yields $F_t^{\mathrm{ref}}$ and $F_t^{\mathrm{gen}}$, respectively. We then measure the discrepancy between the reference and generated ILD distributions at each temporal window using the first Wasserstein distance:
\begin{equation}
d_t^{\mathrm{ILD}}
=
W_1
\left(
F_t^{\mathrm{ref}},
F_t^{\mathrm{gen}}
\right)
=
\int_{-\infty}^{+\infty}
\left|
F_t^{\mathrm{ref}}(x)
-
F_t^{\mathrm{gen}}(x)
\right|
\,\mathrm{d}x.
\end{equation}

Finally, ILD-W for each sample is obtained by averaging the time-aligned Wasserstein distances over all valid reference-active windows:
\begin{equation}
\mathrm{ILD\mbox{-}W}
=
\frac{1}{|\mathcal{T}|}
\sum_{t\in\mathcal{T}}
d_t^{\mathrm{ILD}},
\end{equation}
where $\mathcal{T}$ denotes the set of valid reference-active windows. A window is considered reference-active when the reference audio contains sufficient energy within the analyzed frequency band, ensuring that the ILD distribution is computed only from meaningful acoustic events. Since the Wasserstein distance is computed over the ILD distribution support, ILD-W is reported in decibels. A lower value indicates greater consistency between the generated and reference interaural level-difference distributions over time.

\paragraph{Sound Event Localization and Detection Accuracy (SELD-Acc).}
We evaluate whether the spatial location encoded in the generated stereo audio aligns with the visible sound source using the stereo Sound Event Localization and Detection (SELD) model from SAVGBench~\citep{shimada2026savgbench}. Given a generated stereo waveform, the SELD model predicts the horizontal direction of the dominant sound source at $10\,\mathrm{Hz}$. Each prediction is represented as a normalized horizontal coordinate $x_t^a\in[0,1]$. Since the visual source occupies a spatial region rather than a single point, we represent the visible sound source at time $t$ using the normalized horizontal interval of its bounding box:

\begin{equation}
I_t^v=[x_{0,t}^v,x_{1,t}^v].
\end{equation}

Following SAVGBench, we extend the predicted horizontal position into an interval with a tolerance of $h$ in the original image resolution.

\begin{equation}
I_t^a
=
\left[
x_t^a-h,
x_t^a+h
\right],
\end{equation}

For each reference-active window, the localization is considered correct if the predicted audio interval overlaps with the visible sound-source interval. Otherwise, the localization is considered incorrect, including cases where the SELD model produces no valid prediction. The final SELD-Acc is computed as the ratio of correctly localized windows:

\begin{equation}
\mathrm{SELD\mbox{-}Acc}
=
\frac{N_{\mathrm{correct}}}
{N_{\mathrm{active}}},
\end{equation}

where $N_{\mathrm{correct}}$ denotes the number of reference-active windows with successful audiovisual spatial matching, and $N_{\mathrm{active}}$ is the total number of reference-active windows. A higher SELD-Acc indicates stronger spatial consistency between the generated stereo audio and the visible sound source. Additionally, We note that the SELD model adopted from SAVGBench is primarily validated on speech and musical-instrument sounds, while its localization reliability for other sound categories is less established. Therefore, its predictions should not be interpreted as a universally reliable localization measure across all sound categories in SWBench. Nevertheless, we retain SELD-Acc as a complementary metric because it provides an existing model-based measure of audiovisual spatial alignment, and a substantial portion of SWBench consists of speech-related samples that fall within the evaluation domain considered by SAVGBench. We therefore interpret SELD-Acc together with the other spatial metrics.

\paragraph{Stereo Magnitude Ratio Error (SMR-Err).}
SMR-Err evaluates whether the generated stereo audio preserves the relative amount of differential stereo energy present in the reference audio. Specifically, the mid component captures the common signal shared by the left and right channels, which primarily represents the underlying acoustic content, while the side component captures their differential signal, which primarily reflects stereo spatial effects and inter-channel spatial variation. For each temporal window $t$, we transform the left and right channels into mid and side components:

\begin{equation}
M_t(n)
=
\frac{L_t(n)+R_t(n)}{2},
\qquad
S_t(n)
=
\frac{L_t(n)-R_t(n)}{2}.
\end{equation}

Their corresponding energies are computed as
\begin{equation}
E_t^{M}
=
\frac{1}{N_t}
\sum_{n=1}^{N_t}
M_t(n)^2,
\qquad
E_t^{S}
=
\frac{1}{N_t}
\sum_{n=1}^{N_t}
S_t(n)^2.
\end{equation}

The stereo magnitude ratio is then defined in the logarithmic domain as
\begin{equation}
\mathrm{SMR}_t
=
10\log_{10}
\left(
\frac{E_t^{S}+\epsilon}
{E_t^{M}+\epsilon}
\right).
\end{equation}

A larger side component indicates stronger inter-channel differences, whereas duplicated or nearly mono audio produces a very small side-to-mid ratio. We compare the generated and reference ratios at each reference-active window and define
\begin{equation}
\mathrm{SMR\mbox{-}Err}
=
\frac{1}{|\mathcal{T}|}
\sum_{t\in\mathcal{T}}
\left|
\mathrm{SMR}_t^{\mathrm{gen}}
-
\mathrm{SMR}_t^{\mathrm{ref}}
\right|,
\end{equation}

where $\mathcal{T}$ denotes the set of valid reference-active windows. SMR-Err is measured in decibels, and a lower value indicates that the generated audio more faithfully preserves the stereo spatial balance of the reference audio.

\paragraph{Spatial-AST Angular Consistency (AST-Ang).}
We further employ the pretrained Spatial-AST model~\citep{zheng2024bat} to evaluate spatial correspondence in a learned representation space. Following its evaluation protocol, both reference and generated stereo audio are resampled to $32\,\mathrm{kHz}$ and padded or truncated to a fixed duration before being fed into Spatial-AST. The model produces dedicated representations for source direction and distance. Let $\mathbf{z}_{\mathrm{ang}}^{\mathrm{ref}}$ and $\mathbf{z}_{\mathrm{ang}}^{\mathrm{gen}}$ denote the direction-of-arrival representations extracted from the reference and generated audio, respectively. We define Spatial-AST Angular Consistency as their cosine similarity:
\begin{equation}
\mathrm{AST\mbox{-}Ang}
=
\frac{
\left\langle
\mathbf{z}_{\mathrm{ang}}^{\mathrm{ref}},
\mathbf{z}_{\mathrm{ang}}^{\mathrm{gen}}
\right\rangle
}{
\left\|
\mathbf{z}_{\mathrm{ang}}^{\mathrm{ref}}
\right\|_2
\left\|
\mathbf{z}_{\mathrm{ang}}^{\mathrm{gen}}
\right\|_2
}.
\end{equation}

A higher AST-Ang indicates stronger agreement between the reference and generated audio in the learned spatial-direction representation.

\paragraph{Spatial-AST Calibration (AST-Cal).}

Raw Spatial-AST similarity may be influenced by both the acoustic content and the spatial cues. To evaluate the contribution of spatial information alone, we further construct a duplicated-mono version of each generated sample as a non-spatial baseline:

\begin{equation}
A^{\mathrm{mono}}
=
\left[
\frac{L^{\mathrm{gen}}+R^{\mathrm{gen}}}{2},
\frac{L^{\mathrm{gen}}+R^{\mathrm{gen}}}{2}
\right].
\end{equation}

We extract both angular and distance representations from the generated stereo signal and its duplicated-mono counterpart. For $q\in\{\mathrm{ang},\mathrm{dis}\}$, let
\begin{equation}
s_q
=
\mathrm{cos}
\left(
\mathbf{z}_q^{\mathrm{ref}},
\mathbf{z}_q^{\mathrm{gen}}
\right),
\qquad
b_q
=
\mathrm{cos}
\left(
\mathbf{z}_q^{\mathrm{ref}},
\mathbf{z}_q^{\mathrm{mono}}
\right),
\end{equation}
where $s_q$ denotes the stereo similarity and $b_q$ denotes the corresponding duplicated-mono baseline. We calibrate each similarity relative to this baseline as
\begin{equation}
c_q
=
\mathrm{clip}
\left(
\frac{s_q-b_q}
{\max(1-b_q,\epsilon)},
-1,
1
\right).
\end{equation}

The final calibrated score combines the angular and distance components:
\begin{equation}
\mathrm{AST\mbox{-}Cal}
=
\frac{1}{2}
\left(
c_{\mathrm{ang}}
+
c_{\mathrm{dis}}
\right).
\end{equation}

This calibration measures the improvement of the generated stereo representation over its spatially collapsed mono counterpart. A score of $1$ corresponds to perfect agreement with the reference representation, while $0$ indicates no improvement over the duplicated-mono baseline. Higher AST-Cal therefore indicates stronger preservation of spatial information that specifically arises from the stereo structure.

\section{Analysis of Visual Motion–Track Direction Inconsistency}

To examine how StereoBind exploits different spatial conditions, we conduct a controlled track-reversal experiment that introduces an explicit conflict between the numerical track and its visual motion reference. Given an input track $P\in\mathbb{R}^{121\times2}$, we render its sparse motion reference $R_P$ and reverse only the numerical track $P$ to obtain $P_{\mathrm{rev}}$, while keeping $R_P$ unchanged. The resulting pair $(P_{\mathrm{rev}},R_P)$ therefore encodes opposite motion directions and is compared against the standard spatially consistent setting. As shown in Table~\ref{tab:reverse_track}, introducing this conflict consistently degrades both signal-level and learned spatial metrics. This confirms that StereoBind is sensitive to the consistency of its spatial conditions. Notably, the degradation remains moderate rather than causing complete stereo collapse. The audiovisual backbone and learned stereo prior can still maintain inter-channel variation. Overall, the track-reversal intervention shows that generating stereo differences alone is insufficient for dynamic spatial correspondence. The latter additionally requires those inter-channel differences to evolve consistently with the motion of the visible sound source.

\begin{table}[t]
\centering
\caption{
\textbf{Conflicting spatial condition analysis on SWBench.}
We reverse the numerical motion track while keeping its sparse visual motion reference unchanged, creating contradictory spatial conditions.
$\uparrow$ indicates higher is better, while $\downarrow$ indicates lower is better.
}
\label{tab:reverse_track}
\scriptsize
\setlength{\tabcolsep}{4pt}
\renewcommand{\arraystretch}{1.15}

\begin{tabular}{lccccc}
\toprule
\textbf{Setting}
& \makecell{\textbf{ILD}\\\textbf{-W} $\downarrow$}
& \makecell{\textbf{SELD}\\\textbf{-Acc} $\uparrow$}
& \makecell{\textbf{SMR}\\\textbf{-Err} $\downarrow$}
& \makecell{\textbf{AST}\\\textbf{-Ang} $\uparrow$}
& \makecell{\textbf{AST}\\\textbf{-Cal} $\uparrow$} \\
\midrule

Consistent Spatial Conditions
& \textbf{2.754}
& \textbf{0.757}
& \textbf{14.86}
& \textbf{0.465}
& \textbf{0.314} \\

Reversed Track Conflict
& 2.925
& 0.651
& 20.78
& 0.416
& 0.322 \\

\bottomrule
\end{tabular}
\end{table}

\section{Future Work}

Future work will extend stereoscopic audiovisual generation toward more complex and interactive settings. A key direction is multi-source stereo audio generation, where multiple visible sound sources must be jointly associated with distinct acoustic events and spatial trajectories. Beyond conventional generation, dynamic audiovisual correspondence could also be integrated into world models to support joint prediction of scene evolution, physical interactions, and viewpoint-dependent acoustic responses. Improving generalization to unconstrained real-world scenes, including camera motion, occlusion, off-screen sources, rapid dynamics, and reverberant environments, remains another important challenge.

\end{document}